\documentclass[10pt,twocolumn,letterpaper]{article}

\usepackage{cvpr}
\usepackage{times}
\usepackage{epsfig}
\usepackage{graphicx}
\usepackage{amsmath}
\usepackage{amssymb}
\usepackage{bm}
\usepackage{multirow}
\usepackage{algorithm}
\usepackage{algpseudocode}

\newcommand{\image}[1]{\raisebox{-.75\height}{\includegraphics[width=1.2cm, height=1.2cm]{#1}}}
\newcommand{\triple}[1]{\image{fig/fig6/#1_inputs.jpg}&\image{fig/fig6/#1_targets.jpg}&\image{fig/fig6/#1_notext.jpg}& \image{fig/fig6/#1_text.jpg}&}
\newcommand{\figsixtable}[2]{\triple{#1} #2 \\ \hline} 
\newcommand{\triplesupp}[1]{\image{fig/fig_supp1/#1_inputs.jpg}&\image{fig/fig_supp1/#1_targets.jpg}&\image{fig/fig_supp1/#1_i2i.jpg}&\image{fig/fig_supp1/#1_lbie.jpg}&}
\newcommand{\figsupptable}[2]{\triplesupp{#1} #2 \\ \hline}

\usepackage[pagebackref=true,breaklinks=true,letterpaper=true,colorlinks,bookmarks=false]{hyperref}

\cvprfinalcopy 

\def\cvprPaperID{1956} 

\ifcvprfinal\fi
\begin{document}

\title{Language-Based Image Editing with Recurrent Attentive Models}

\author{
  Jianbo Chen$^{*}$, Yelong Shen$^{\dagger}$, Jianfeng Gao$^{\dagger}$, Jingjing Liu$^{\dagger}$, Xiaodong Liu$^{\dagger}$\\
  University of California, Berkeley$^*$ and Microsoft Research$^\dagger$\\
  \texttt{jianbochen@berkeley.edu}\\
  \texttt{yeshen, jfgao, jingjl, xiaodl@microsoft.com}\\
}

\maketitle

\begin{abstract}
   We investigate the problem of Language-Based Image Editing (LBIE). Given a source image and a natural language description, we want to generate a target image by editing the source image based on the description. We propose a generic modeling framework for two sub-tasks of LBIE: language-based image segmentation and image colorization. The framework uses recurrent attentive models to fuse image and language features. Instead of using a fixed step size, we introduce for each region of the image a termination gate to dynamically determine after each inference step whether to continue extrapolating additional information from the textual description. The effectiveness of the framework is validated on three datasets. First, we introduce a synthetic dataset, called CoSaL, to evaluate the end-to-end performance of our LBIE system. Second, we show that the framework leads to state-of-the-art performance on image segmentation on the ReferIt dataset. Third, we present the first language-based colorization result on the Oxford-102 Flowers dataset.
\end{abstract}

\section{Introduction} 
\begin{figure}[t]
\centering
\includegraphics[width=0.43\linewidth]{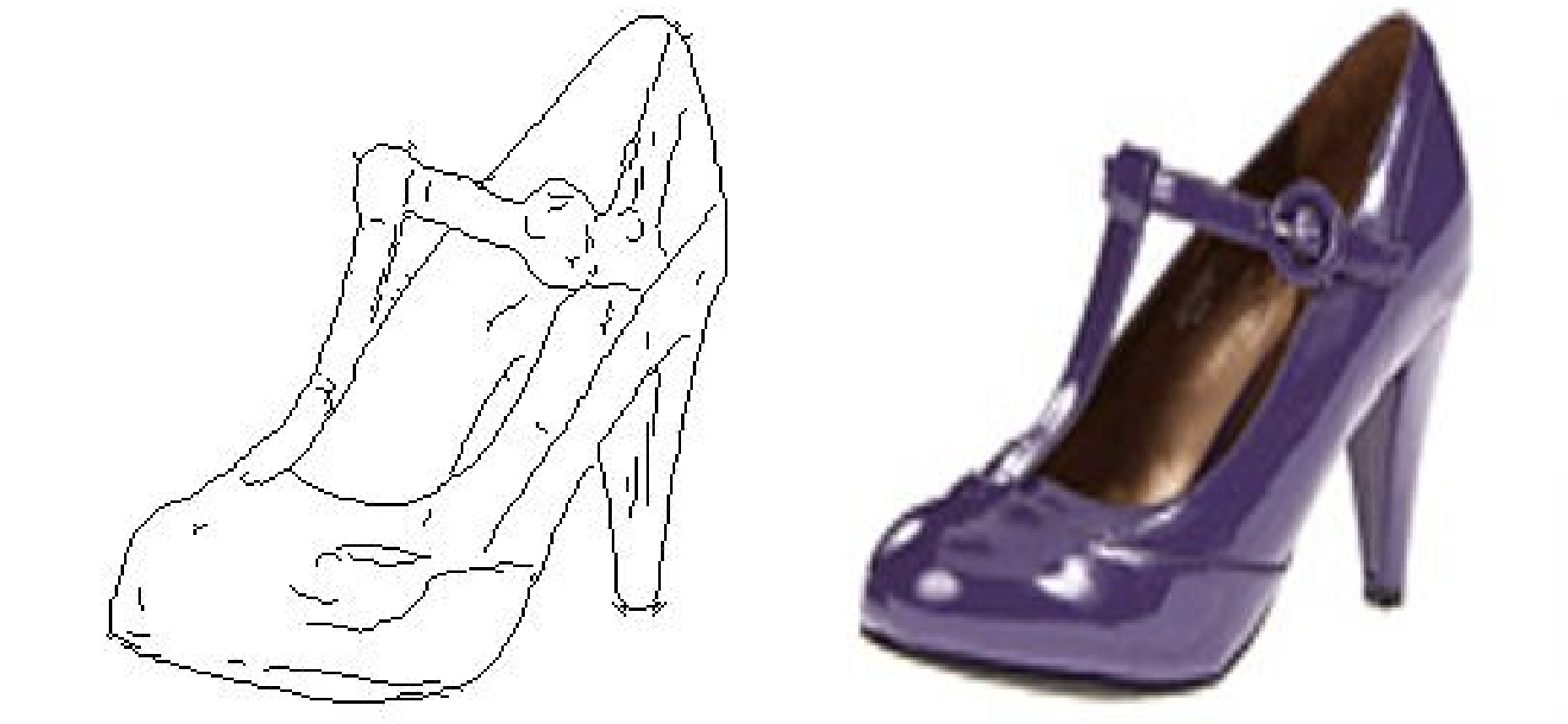} 
\caption{In an interactive design interface, a sketch of shoes is presented to a customer, who then gives a verbal instruction on how to modify the design: ``\textit{The insole of the shoes should be brown. The vamp and the heel should be purple and shining}''. The system colorizes the sketch following the customer's instruction. (images from \cite{isola2016image}).}
\label{fig:design} 
\end{figure}

\begin{figure}[t]
\centering
\includegraphics[width=0.43\linewidth]{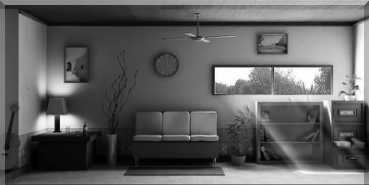}
\includegraphics[width=0.43\linewidth]{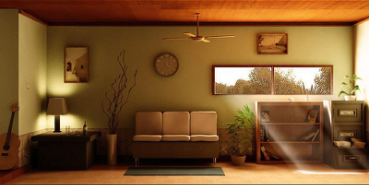}    
\caption{The image on the left is an initial virtual environment. The user provides a textual description: ``The afternoon light flooded the little room from the window, shining the ground in front of a brown bookshelf made of wood. Besides the bookshelf lies a sofa with light-colored cushions. There is a blue carpet in front of the sofa, and a clock with dark contours above it...''. The system modifies the virtual environment into the target image on the right.}
\label{fig:VR} 
\end{figure}

\vspace{-1mm}
In this work, we aim to develop an automatic Language-Based Image Editing (LBIE) system. Given a source image, which can be a sketch, a grayscale image or a natural image, the system will automatically generate a target image by editing the source image following natural language instructions provided by users. Such a system has a wide range of applications from Computer-Aided Design (CAD) to Virtual Reality (VR). As illustrated in Figure \ref{fig:design}, a fashion designer presents a sketch of a pair of new shoes (i.e., the source image) to a customer, who can provide modifications on the style and color in verbal description, which can then be taken by the LBIE system to change the original design. The final output (i.e., the target image) is the revised and enriched design that meets the customer’s requirement. Figure \ref{fig:VR} showcases the use of LBIE for VR. While most VR systems still use button-controlled or touchscreen interface, LBIE provides a natural user interface for future VR systems, where users can easily modify the virtual environment via natural language instructions.
 
\begin{figure}[t]
\centering
\includegraphics[width=0.33\linewidth, height = 1.8cm]{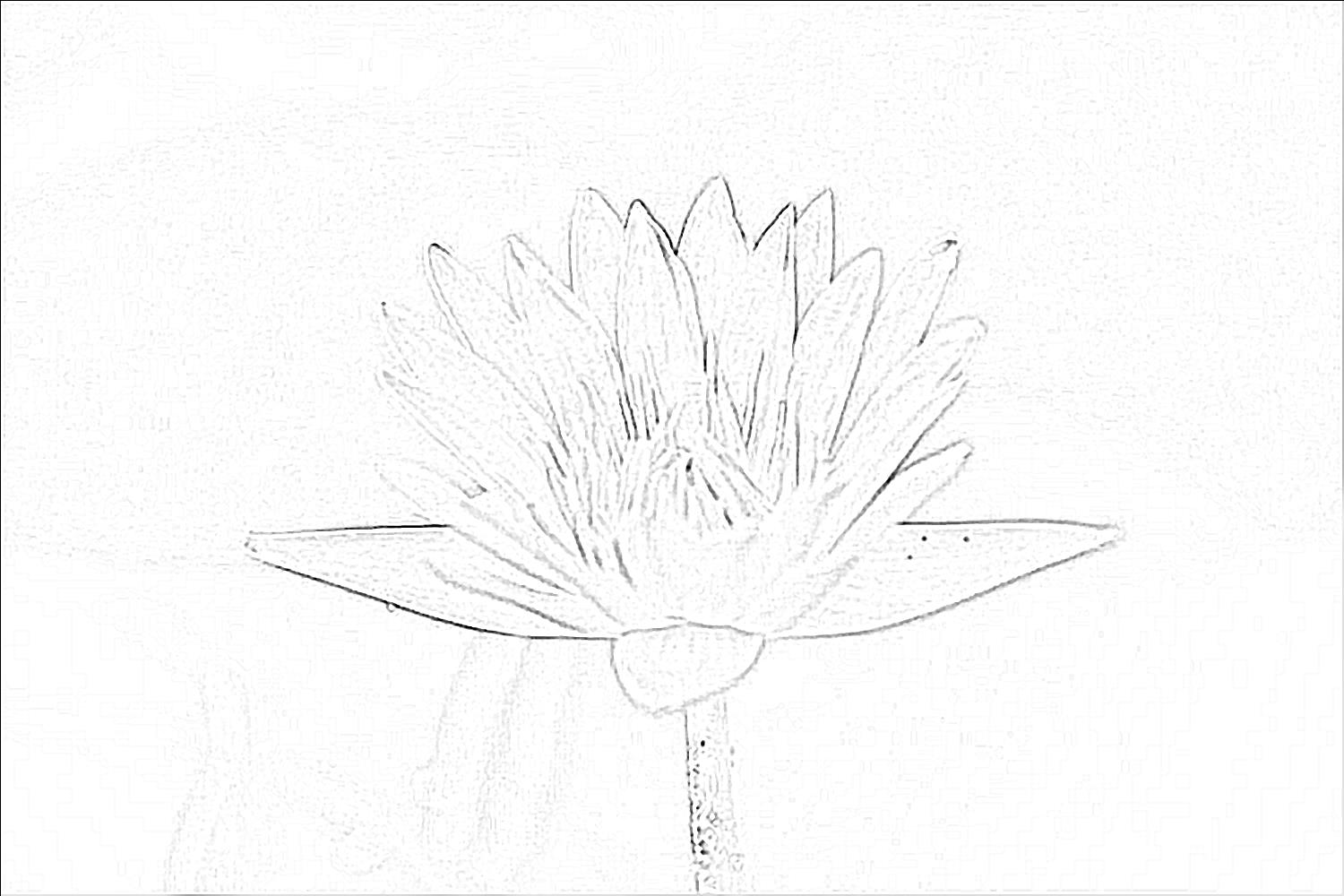}%
\includegraphics[width=0.33\linewidth, height = 1.8cm]{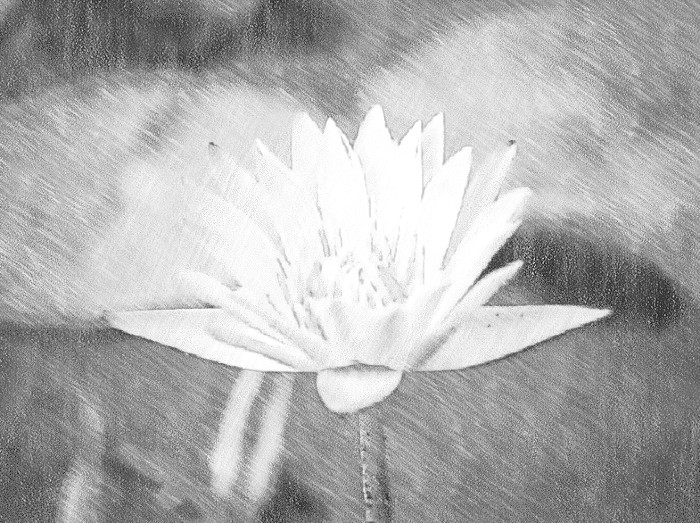} 
\includegraphics[width=0.33\linewidth, height = 1.8cm]{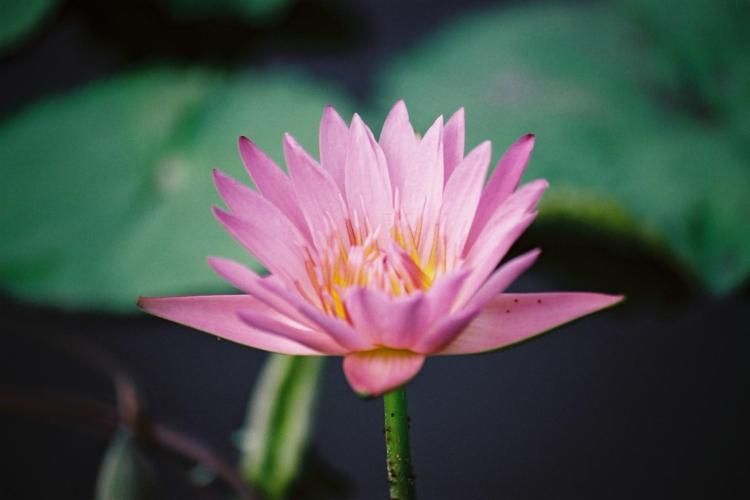}  
\caption{Left: sketch image. Middle: grayscale image. Right: color image (from \cite{Nilsback08}). A language-based image editing system will take either of the first two images as the input, and generate the third color image following a natural language expression: \textit{``The flower has red petals with yellow stigmas in the middle''},.}
\label{fig:rendering} 
\vspace{-4mm}
\end{figure}
LBIE covers a broad range of tasks in image generation: shape, color, size, texture, position, etc. This paper focuses on two basic sub-tasks: language-based segmentation and colorization for shapes and colors. As shown in Figure \ref{fig:rendering}, given a grayscale image and the expression \textit{``The flower has red petals with yellow stigmas in the middle''}, the segmentation model will identify regions of the image as ``\textit{petals}'', ``\textit{stigmas}'', and the colorization model will paint each pixel with the suggested color. 
In this intertwined task of segmentation and colorization, the distribution of target images can be multi-modal in the sense that each pixel will have a definitive ground truth on segmentation, but not necessarily on color. For example, the pixels on petals in Figure \ref{fig:rendering} should be red based on the textual description, but the specific numeric values of the red color in the RGB space is not uniquely specified. The system is required to colorize the petals based on real-world knowledge. Another uncertainty lies in the fact that the input description might not cover every detail of the image. The regions that are not described, such as the leaves in the given example, need to be rendered based on common sense knowledge. In summary, we aim to generate images that not only are consistent with the natural language expressions, but also align with common sense.

Language-based image segmentation has been studied previously in \cite{hu2016segmentation}. However, our task is far more challenging because the textual description often contains multiple sentences (as in Figure \ref{fig:VR}), while in \cite{hu2016segmentation} most of the expressions are simple phrases. To the best of our knowledge, language-based colorization has not been studied systematically before. In most previous work, images are generated either solely based on natural language expressions \cite{reed2016generative},\cite{zhang2016stackgan} or based on another image \cite{isola2016image},\cite{cheng2015deep},\cite{zhang2016colorful}. Instead, we want to generate a target image based on both the natural language expression and the source image. Related tasks will be discussed in detail in Section \ref{sec:related}.

A unique challenge in language-based image editing is the complexity of natural language expressions and their correlation with the source images. As shown in Figure \ref{fig:VR}, the description usually consists of multiple sentences, each referring to multiple objects in the source image. When human edits the source image based on a textual description, we often keep in mind which sentences are related to which region/object in the image, and go back to the description multiple times while editing that region. This behavior of ``going back'' often varies from region to region, depending on the complexity of the description for that region. An investigation of this problem is carried out on CoSaL, which is a synthetic dataset described in Section \ref{sec:experiment}.

\begin{figure*}[t]
\centering
\includegraphics[width=0.8\textwidth,height=7cm]{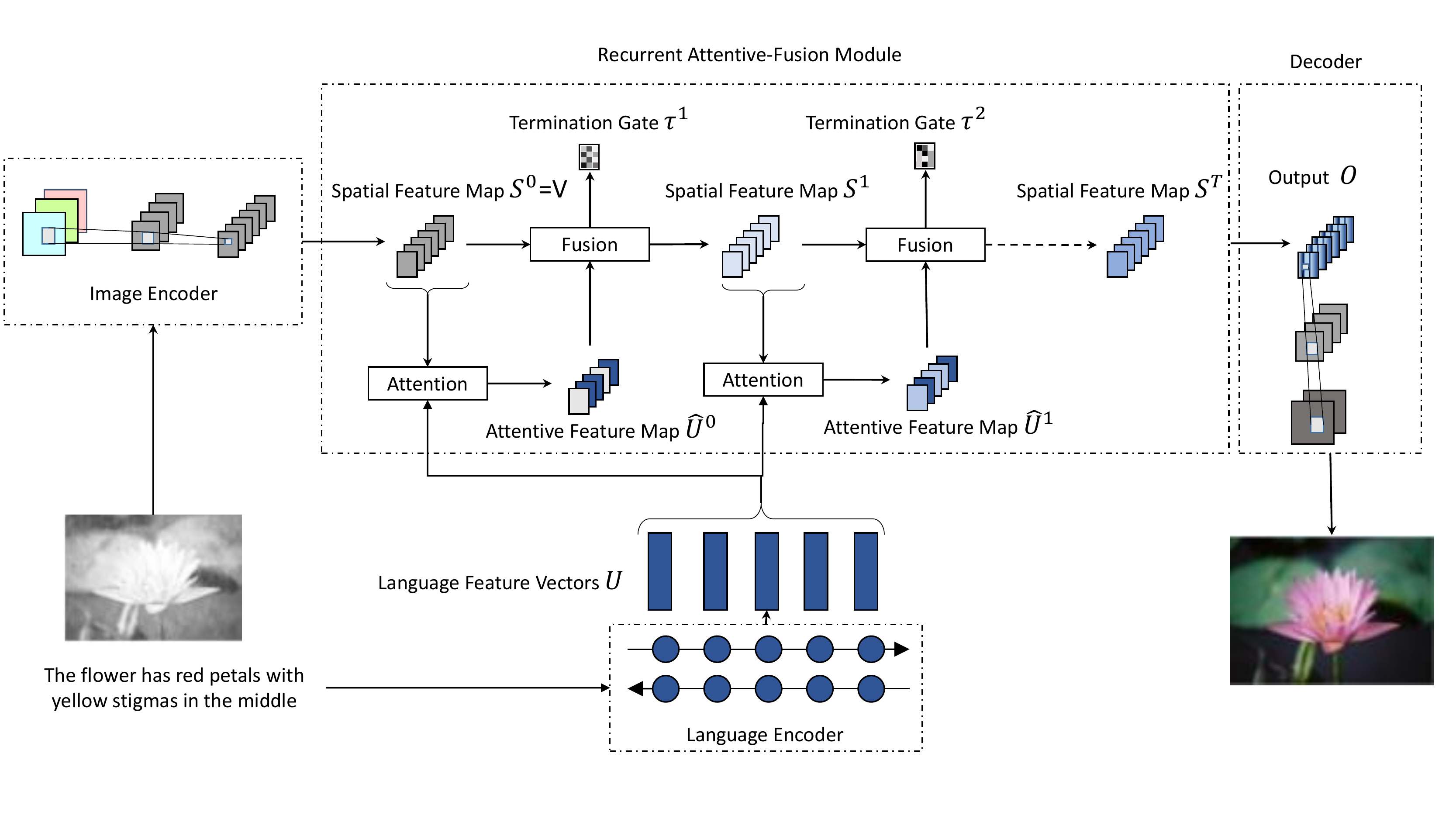} 
\caption{A high-level diagram of our model, composed of a convolutional image encoder, an LSTM text encoder, a fusion module, a deconvolutional upsampling layer, with an optional convolutional discriminator.}
\label{fig:model} 
\vspace{-2mm}
\end{figure*}

Our goal is to design a generic framework for the two sub-tasks in language-based image editing. A diagram of our model is shown in Figure \ref{fig:model}. Inspired by the observation aforementioned, we introduce a recurrent attentive fusion module in our framework. The fusion module takes as input the image features that encode the source image via a convolutional neural network, and the textual features that encode the natural language expression via an LSTM, and outputs the fused features to be upsampled by a deconvolutional network into the target image. In the fusion module, recurrent attentive models are employed to extract distinct textual features based on the spatial features from different regions of an image. A termination gate is introduced for each region to control the number of steps it interacts with the textual features. The Gumbel-Softmax reparametrization trick \cite{jang2016categorical} is used for end-to-end training of the entire network. Details of the models and the training process are described in Section \ref{sec:model}.

Our contributions are summarized as follows:
\begin{itemize}
	\setlength\itemsep{0.001em}
	\item We define a new task of language-based image editing (LBIE).
	\item We present a generic modeling framework based on recurrent attentive models for two sub-tasks of LBIE: language-based image segmentation and colorization.
	\item We introduce a synthetic dataset CoSaL designed specifically for the LBIE task. 
	\item We achieve new state-of-the-art performance on language-based image segmentation on the ReferIt dataset.
	\item We present the first language-based colorization result on the Oxford-102 Flowers dataset, with human evaluations validating the performance of our model.
	\vspace{-2mm} 
\end{itemize}

\vspace{-2mm} 
\section{Related Work} \label{sec:related}
\vspace{-1mm}
While the task of language-based image editing has not been studied, the community has taken significant steps in several related areas, including Language Based object detection and Segmentation (LBS) \cite{hu2016segmentation},\cite{hu2016natural}, Image-to-Image Translation (IIT) \cite{isola2016image}, Generating Images from Text (GIT) \cite{reed2016learning}, \cite{zhang2016stackgan}, Image Captioning (IC) \cite{karpathy2015deep}, \cite{vinyals2015show}, \cite{xu2015show}, Visual Question Answering (VQA) \cite{antol2015vqa}, \cite{yang2016stacked},  Machine Reading Comprehension (MRC) \cite{hermann2015teaching}, etc. We summarize the types of inputs and outputs for these related tasks in Table \ref{table:tasks}.  

\begin{table}[h!]
\centering
\begin{tabular}{||c| c c| c c||} 
\hline
& \multicolumn{2}{c}{Inputs}&\multicolumn{2}{c||}{Outputs}  \\ 
 \hline 
 & Text &Image&Text&Image \\ 
 \hline 
 MRC &YES&NO&YES&NO \\
 VQA & YES &YES&YES&NO \\ 
 IIT & NO &YES&NO&YES \\ 
 IC & NO &YES&YES&NO \\ 
 GIT& YES &NO&NO&YES \\ 
 LBS & YES &YES&NO&YES \\ 
 \textbf{LBIE} & \textbf{YES} &\textbf{YES}&\textbf{NO}&\textbf{YES} \\ 
 \hline
\end{tabular}
\caption{The types of inputs and outputs for related tasks}
\label{table:tasks}
\end{table}
\vspace{-2mm}
\subsection*{Recurrent attentive models}
\vspace{-1mm}
Recurrent attentive models have been applied to visual question answering (VQA) 
to fuse language and image features \cite{yang2016stacked}. The stacked attention network proposed in \cite{yang2016stacked} identifies the image regions that are relevant to the question via multiple attention layers, which can progressively filter out noises and pinpoint the regions relevant to the answer. In image generation, a sequential variational auto-encoder framework, such as DRAW\cite{gregor2015draw}, has shown substantial improvement over standard variational auto-encoders (VAE) \cite{kingma2013auto}. Similar ideas have also been explored for machine reading comprehension, where models can take multiple iterations to infer an answer based on the given query and document \cite{dhingra2016gated}, \cite{weston2015towards}, \cite{wang2016machine}, \cite{xiong2016dynamic}, \cite{liu2017stochastic}. In \cite{shen2017reasonet} and \cite{shen2016implicit}, a novel neural network architecture called ReasoNet is proposed for reading comprehension. ReasoNet performs multi-step inference where the number of steps is determined by a termination gate according to the difficulty of the problem. ReasoNet is trained using policy gradient methods.

\vspace{-2mm}
\subsection*{Segmentation from language expressions}
\vspace{-1mm}
The task of language-based image segmentation is first proposed in \cite{hu2016segmentation}. Given an image and a natural language description, the system will identify the regions of the image that correspond to the visual entities described in the text. The authors in \cite{hu2016segmentation} proposed an end-to-end approach that uses three neural networks: a convolutional network to encode source images, an LSTM network to encode natural language descriptions, and a fully convolutional classification and upsampling network for pixel-wise segmentation. 

One of the key differences between their approach and ours is the way of integrating image and text features. In \cite{hu2016segmentation}, for each region in the image, the extracted spatial features are concatenated with the same textual features. 
Inspired by the alignment model of \cite{karpathy2015deep}, in our approach, each spatial feature is aligned with different textual features based on attention models. Our approach yields superior segmentation results than that of \cite{hu2016segmentation} on a benchmark dataset.
\vspace{-2mm}
\subsection*{Conditional GANs in image generation}
\vspace{-1mm}
Generative adversarial networks (GANs) \cite{goodfellow2014generative} have been widely used for image generation. Conditional GANs \cite{mirza2014conditional} are often employed when there are constraints that a generated image needs to satisfy. For example, deep convolutional conditional GANs \cite{radford2015unsupervised} have been used to synthesize images based on textual descriptions \cite{reed2016generative} 
\cite{zhang2016stackgan}. \cite{isola2016image} proposed the use of conditional GANs for image-to-image translation.
Different from these tasks, LBIE takes both image and text as input, presenting an additional challenge of fusing the features of the source image and the textual description.
\vspace{-2mm}
\section{The Framework}\label{sec:model}
\paragraph{Overview} The proposed modeling framework, as shown in \ref{fig:model}, is based on neural networks, and is generic to both the language-based image segmentation and colorization tasks. The framework is composed of a convolutional image encoder, an LSTM text encoder, a fusion network that generates a fusion feature map by integrating image and text features, a deconvolutional network that generates pixel-wise outputs (the target image) by upsampling the fusion feature map, and an optional convolutional discriminator used for training colorization models.

\vspace{-5mm}
\paragraph{Image encoder} The image encoder is a multi-layer convolutional neural network (CNN). Given a source image of size $H\times W$, the CNN encoder produces a $M \times N$ 
spatial feature map, with each position on the feature map containing a $D$-dimensional feature vector ($D$ channels), $V=\{v_{i}:i=1,\dots,M\times N\},v_i\in \mathbb R^{D}.$
\vspace{-5mm}
\paragraph{Language encoder}  The language encoder is a recurrent Long Short-Term Memory (LSTM) network. Given a natural language expression of length $L$, we first embed each word into a vector through a word embedding matrix, then use LSTM to produce for each word a contextual vector that encodes its contextual information such as word order and word-word dependencies. The resulting language feature map is $U=\{u_{i}:i=1,\dots,L\}, u_{i} \in\mathbb R^{K}$. 

\vspace{-5mm}
\paragraph{Recurrent attentive fusion module} The fusion network fuses text information in $U$ into the $M\times N$ image feature map $V$, and outputs an $M\times N$ fusion feature map, with each position (image region) containing an editing feature vector, $O=\{o_i:i=1,\dots,M\times N\}, o_i \in\mathbb  R^{D}$.

The fusion network is devised to mimic the human image editing process. For each region in the source image $v_{i}$, the fusion network reads the language feature map $U$ repeatedly with attention on different parts each time until enough editing information is collected to generate the target image region. The number of steps varies from region to region. 

\vspace{-5mm}
\subparagraph{Internal state} The internal state at time step $t$ is denoted as $S^t=\{s_{i}^t,i=1,\dots,M\times N\}$, which is a spatial feature map, with each poition (image region) containing a vector representation of the editing information state. The initial state is the spatial feature map from the source image, $S^0=V$. The sequence of internal states is modeled by Convolutional Gated Recurrent Units (C-GRUs) which will be described below.
\vspace{-5mm}
\subparagraph{Attention} The attention at time step $t$ is denoted as $\hat{U}^t=\{\hat{u}_{i}^t,i=1,\dots,M\times N\}$, which is a spatial feature map generated based on the current internal state $S^t$ and the language feature map $U$:
\vspace{-2mm} 
\begin{equation*}
\hat{U}^t = \textbf{Attention}(U, S^t; \theta_a), 
\vspace{-2mm}
\end{equation*} 
where \textbf{Attention}(.) is implemented as follows:
\vspace{-2mm}
\begin{align*}
\beta_{ij} &\propto \exp\{{s_{i}^t}^TWu_j\},\\
\hat u_{i}^t &= \sum_{j=1}^L\beta_{ij}u_j.\\
\vspace{-2mm}
\end{align*}

\vspace{-5mm}
\subparagraph{C-GRUs} C-GRUs update the current internal state $S^t$ by infusing the attention feature map $\hat{U}^t$:
\vspace{-2mm}
\begin{equation*}
S^{t+1}=\textbf{C-GRUs}(S^t, \hat{U}^t; \theta_c).
\vspace{-2mm}
\end{equation*}
The \textbf{C-GRUs}(.) is implemented as follows:
\vspace{-2mm}
\begin{align*}
\mathbf z&= \sigma (W_1\otimes S^t + W_2\otimes\hat U^t+b_1),\\
\mathbf r&= \sigma (W_3\otimes S^t + W_4\otimes\hat U^t+b_2),\\
\mathbf c&=\text{ReLU}(W_5\otimes(\mathbf r\odot S^t)+W_6\otimes\hat U^t + b),\\
\hat{O}^t&=\mathbf{h} = (1 - \mathbf z)\odot S^t + \mathbf z\odot \mathbf c,\\
S^{t+1}&= W_7 \otimes \mathbf{h},
\vspace{-2mm}
\end{align*}
where $\odot$ is the elementwise-product, and $\otimes$ is the convolutional operator. Note that $\hat{O}^t$ is the intermediate output of the fusion feature map at time step $t$.


\vspace{-5mm}
\subparagraph{Termination gates} There are $M\times N$ termination gates, each for one image region $v_i$ in $V$. Each termination gate generates a binary random variable according to the current internal state of its image region: $\tau_{i}^t \sim p(\cdot │ f_{tg} (s_{i}^t;\theta_{tg}))$. If $\tau_{i}^t=1$, the fusion process for the image region $v_i$ stops at $t$, and the editing feature vector for this image region is set as $o_{i}=\hat{o}_i^t$. When all terminate gates are true, the fusion process for the entire image is completed, and the fustion network outputs the fusion feature map $O$. \newline
We define $\pmb \zeta = (\zeta_1,\zeta_2,\dots,\zeta_{M\times N})$, where $\zeta_i = (\tau_i^1,\tau_i^2,\dots,\tau_i^T)$, a categorical distribution with $p(\zeta_i=e_t)=\beta_i^t$, where
\vspace{-2mm} 
\begin{equation*} 
\beta_i^t = f_{tg}(s_i^t;\theta_{tg})\prod_{k<t}(1-f_{tg}(s_i^k;\theta_{tg})).
\vspace{-2mm}
\end{equation*}
the probability of stopping the fusion process at the $i$-th image region of the feature map at time $t$. 
\vspace{-5mm}
\subparagraph{Inference} Algorithm \ref{alg:inference} describes the stochastic inference process of the fusion network. The state sequence $S^{(1:T)}$ is hidden and dynamic, chained through attention and C-GRU in a recurrent fashion. The fusion network outputs for each image region $v_i$ an editing feature vector $o_{i}$ at the $t_i$-th step, where $t_i$ is controlled by the $i$th termination gate, which varies from region to region. 
  
  \begin{algorithm}
       \caption{Stochastic Inference of the Fusion Network}
       \label{alg:inference}
       \begin{algorithmic}
       \Require $V\in\mathbb R^{D\times (M\times N)}$: Spatial feature map of image.
       \Require $U\in\mathbb R^{K\times L}$: Language feature map of expression. 
       \Ensure Fusion feature map $O\in\mathbb R^{D\times (M\times N)}$. 
       \Function {Fusion}{$V,U$}
       	  \State Initialize $S^0=V$. 
           \ForAll  {$t=0$ to $t_{max}-1$}
           	\State $\hat U^t = \textbf{Attention}(U,S^t;\theta_a) $ 
 			\State $S^{t+1},\hat{O}^{t} =\textbf{C-GRUs}(S^t,\hat U^t;\theta_c) $
           \State Sample $\pmb{\tau}^{t+1}\sim p(\cdot|f_{tg} (S^{t+1};\theta_{tg}))$ 
           \If {$\tau_{i}^{t+1}=1$ and $\tau_{i}^s=0$ for $s\leq t$}
             \State Set $O_{i}=\hat O_{i}^{t+1}$.
           \EndIf
           \EndFor  
          
           \ForAll {$i=1$ to $M\times N$}
           	\If {$\tau_{i}=0$}
             	\State Set $o_{i}=\hat{o}_{i}^{t_{max}-1} $
             \EndIf
 		  \EndFor	
 	  \EndFunction
       \end{algorithmic}
 \end{algorithm}
\vspace{-5mm}  
\paragraph{Image decoder} The image decoder is a multi-layer deconvolutional network. It takes as input the $M\times N$ fusion feature map $O$ produced by the fusion module, and unsamples from $O$ to produce a $H\times W\times D_e$ editing map $E$ of the same size as the target image, where $D_e$ is the number of classes in segmentation and $2$ ($ab$ channels) in colorization.  
\vspace{-5mm}
\paragraph{Discriminator} The discriminator $D_\phi(E)$ takes in a generated image and its corresponding language description and outputs the probability of the image being realistic. The discriminator uses a convolutional neural network to extract features from the image, as in \cite{reed2016generative}, and uses an LSTM to encode language. The language features are extracted using the attention mechanism and aligned to features extracted from each region of the image respectively. Parameters of the LSTM and the attention map are not shared with those of the previous language encoder. 
\vspace{-5mm}    
\paragraph{Loss and training} 
Denote the loss as $L(\theta)=\mathbb E_{\pmb \zeta}[l(E(\pmb \zeta,\theta),Y)]$, where the expectation is taken over the categorical variables $\pmb \zeta$ generated by termination gates, and $l_\theta(\pmb \zeta)=l(E(\pmb \zeta,\theta),Y)$ is the loss of output at $\pmb \zeta$, and $Y$ is the target image (i.e., the class labels in segmentation or the $ab$ channels in colorization). Denote the probability mass function of $\pmb \zeta$ by $p_\theta(\pmb \zeta)$. Because the sample space is of exponential size $T^{M\times N}$, it is intractable to sum over the entire sample space.  A naive approach to approximation is to subsample the loss and update parameters via the gradient of Monte Carlo estimate of loss:
\vspace{-2mm}
\begin{align*}
\nabla_\theta L(\theta) &= \nabla_\theta\mathbb E_{\pmb \zeta}[l_\theta(\pmb \zeta)]\\
&= \nabla_\theta\mathbb E_{\pmb \zeta}[l_\theta(\pmb \zeta)]\\
&=\sum_{\pmb \zeta}p_\theta (\pmb \zeta) \big ( l_\theta(\pmb \zeta)\nabla_\theta \log p_\theta(\pmb \zeta)\big ) + \nabla_\theta l_\theta(\pmb \zeta)\big )\\
&\approx \frac{1}{|\mathbf S|}\sum_{\pmb \zeta\in\mathbf S} l_\theta(\pmb \zeta)\nabla_\theta \log p_\theta(\pmb \zeta)+\nabla_\theta l_\theta(\pmb \zeta),
\vspace{-2mm}
\end{align*}
where $\mathbf S$ is a subset of $\pmb \zeta$ sampled from the distributon $p_\theta(\pmb \zeta)$. The above update is called a REINFORCE-type algorithm \cite{williams1992simple}. In experiments, we found that the above Monte Carlo estimate suffers from high variance. To resolve this issue, we employ the Gumbel-Softmax reparameterization trick \cite{jang2016categorical}, which replaces every $\zeta_i\in\{0,1\}^T$ sampled from $\text{Cat}(\beta_1,\beta_2,\dots,\beta_T)$ by another random variable $z_i$ generated from Gumbel-Softmax distribution:
\vspace{-2mm}
\begin{equation*}
z_i^t = \frac{\exp((\log\beta_i^t+\varepsilon_i^t)/\lambda)}{\sum_{k=1}^T\exp((\log\beta_i^t+\varepsilon_i^t)/\lambda)},
\vspace{-2mm}
\end{equation*}
where $\lambda$ is a temperature annealed via a fixed schedule and the auxiliary random variables $\varepsilon_i^1,\dots,\varepsilon_i^T$ are i.i.d. samples drawn from Gumbel$(0,1)$ independent of the parameters $\beta_i$:
\vspace{-2mm}
\begin{equation*}
\varepsilon_i^t = -\log(-\log u_i^t), u_i^t\sim \text{Unif}(0,1).
\vspace{-2mm}
\end{equation*}  
Define $\mathbf z(\pmb\varepsilon,\theta) = (z_1,z_2,\dots,z_{MN})$. The loss can be rewritten as $L(\theta) = \mathbb E_{\pmb\varepsilon}[l_\theta(\pmb z(\pmb\varepsilon,\theta) )]$, and the update is approximated by taking the gradient of Monte Carlo estimates of the loss obtained from sampling $\pmb\varepsilon$. 

We use two different losses for segmentation and colorization, respectively.   
\vspace{-5mm}
\subparagraph{Segmentation} In segmentation, we assume there is a unique answer for each pixel on whether or not it is being referred in the stage of segmentation. The response map $E$ is of size $H\times W\times D_e$, which produces a log probability for each class for each pixel. We use a pixel-wise softmax cross-entropy loss during training:
\vspace{-2mm}
\begin{equation*}
l(E,Y) = \text{Cross-Entropy}(\text{Softmax}(E), Y).
\vspace{-2mm}
\end{equation*}
\vspace{-5mm}
\subparagraph{Colorization} In colorization, the high-level goal is to generate realistic images under the constraint of natural language expressions and input scene representations, we introduce a mixture of GAN loss and $L1$ loss for optimization as in \cite{isola2016image}. A discriminator $D_\phi$ parametrized by $\phi$ is introduced for constructing the GAN loss. 

The response map $E$ is the predicted $ab$ color channels. It is combined with the grayscale source image to produce a generated color image $E'$. The generator loss is a GAN loss taking $E'$ as input, and $L1$ loss between the $ab$ channels of the target image $Y$ and the response map $E$:
\vspace{-2mm}
\begin{equation*}
l(E,Y) = \log (1-D_\phi(E)) + \gamma \|E-Y\|_1 \text{ } (\gamma=0.01).
\vspace{-2mm}
\end{equation*}

The discriminator $D_\phi$ is trained by first generating a sample $E$ via Algorithm \ref{alg:inference}, combined with the grayscale source image to produce $E'$, and optimize the following loss over $\phi$:
\vspace{-2mm}
\begin{equation*}
\log (D_\phi(E')) + \log (1-D_\phi(Y)).
\vspace{-2mm}
\end{equation*}

The generator loss and the discriminator loss are optimized alternatively in the training stage.  

\vspace{-2mm}
\section{Experiments}\label{sec:experiment}
We conducted three experiments to validate the performance of the proposed framework. A new synthetic dataset CoSaL (Colorizing Shapes with Artificial Language) was introduced to test the capability of understanding multi-sentence descriptions and associating the inferred textual features with visual features. Our framework also yielded state-of-the-art performance on the benchmark dataset ReferIt \cite{KazemzadehOrdonezMattenBergEMNLP14} for image segmentation. A third experiment was carried out on the Oxford-102 Flowers dataset \cite{Nilsback08}, for the language-based colorization task. All experiments were coded in TensorFlow. Codes for reproducing the key results are available online\footnote{\url{https://github.com/Jianbo-Lab/LBIE}}.
\subsection{Experiments on CoSaL}
\begin{figure}[H]
\centering
\includegraphics[width=0.57\linewidth]{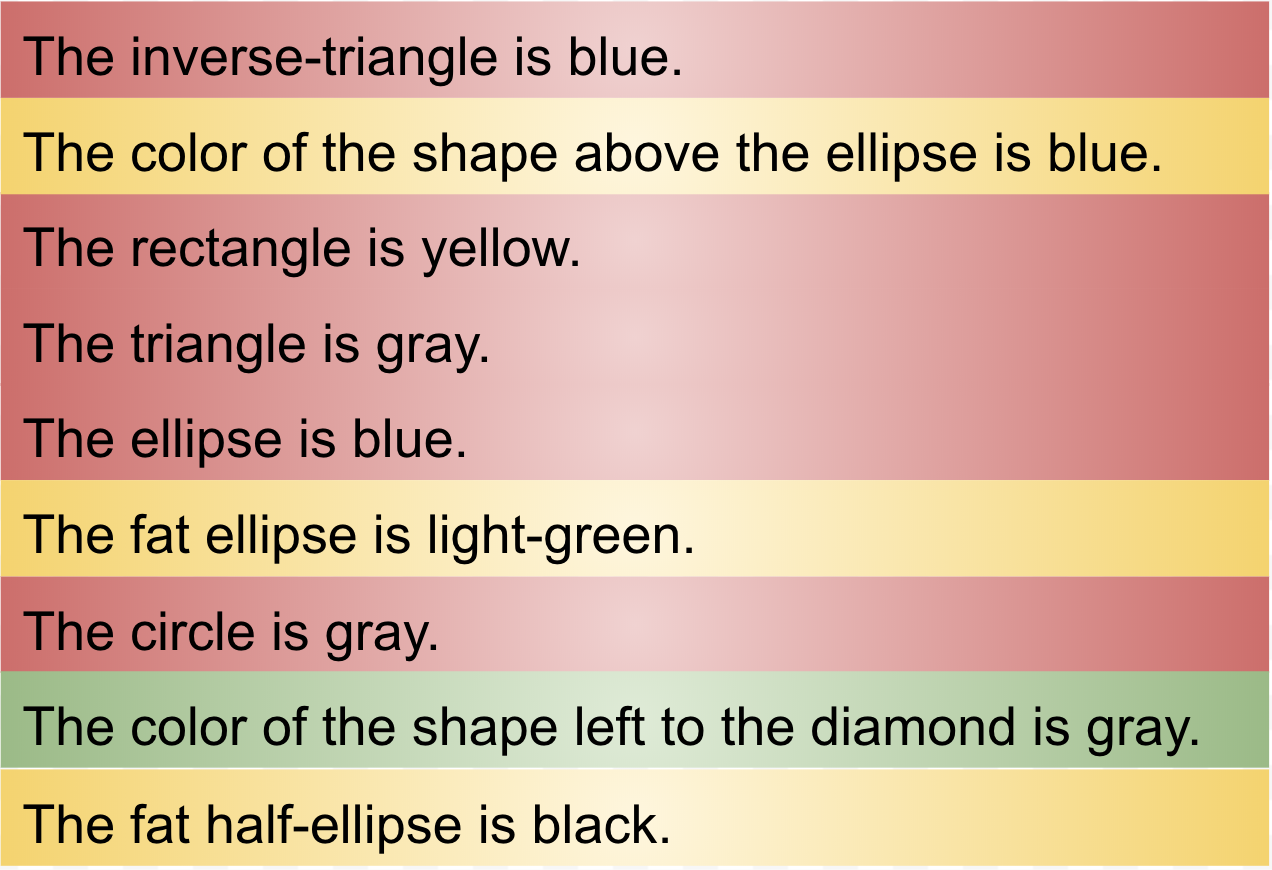}%
\includegraphics[width=0.42\linewidth]{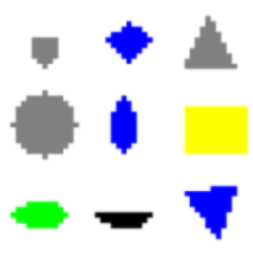}  
\caption{Right: ground truth image. Left: illustration of which sentences are attended to at each time step. Red, yellow and green represent the first, second and third time step, respectively.}
\vspace{-3mm}
\label{fig:cosal2} 
\end{figure}

\vspace{-2mm}
\paragraph{Dataset} 
Each image in the CoSaL dataset consists of nine shapes, paired with a textual description of the image. The task is defined as: given a black-white image and its corresponding description, colorize the nine shapes following the textual description. Figure \ref{fig:cosal2} shows an example. It requires sophisticated coreference resolution, multi-step inference and logical reasoning to accomplish the task. 

The dataset was created as follows: first, we divide a white-background image into $3\times 3$ regions. Each region contains a shape randomly sampled from a set of $S$ shapes (e.g., squares, fat rectangles, tall rectangles, circles, fat ellipses, tall ellipses, diamonds, etc.)  Each shape is then filled with one of $C$ color choices, chosen at random. The position and the size of each shape are generated by uniform random variables. As illustrated in Figure \ref{fig:cosal2}, the difficulty of this task increases with the number of color choices. In our experiments, we specify $C=3$.

The descriptive sentences for each image can be divided into two categories: direct descriptions and relational descriptions. The former prescribes the color of a certain shape (e.g., \textit{Diamond is red}), and the latter depicts one shape conditional of another (e.g., \textit{The shape left to Diamond is blue}). To understand direct descriptions, the model needs to associate a specified shape with its textual features. Relational description adds another degree of difficulty, which requires advanced inference capability of relational/multi-step reasoning. The ratio of direct descriptions to relational descriptions varies among different images, and all the colors and shapes in each image are uniquely determined by the description. In our experiment, we randomly generated $50,000$ images with corresponding descriptions for training purpose, and $10,000$ images with descriptions for testing. 
\vspace{-2mm}
\begin{table}[H]
\centering
\begin{tabular}{||c c c c c||} 
\hline
&&  \multicolumn{3}{c||}{Number of direct descriptions}  \\ 
 \hline

 $T$ & Attention &4&6&8  \\ [0.5ex] 
 \hline\hline
 1 & No &0.2107&0.2499&0.3186 \\
 \hline
 1 & Yes &0.4030&0.5220&\textbf{0.7097} \\
 \hline
 4 & Yes & \textbf{0.5033}&\textbf{0.5313}&0.7017 \\ 
 \hline
\end{tabular}
\caption{The average IoU of two models, without attention at $T=1$ and with attention at $T=1,4$. Performance varies among datasets with different ratios of direct to relational descriptions.}
\vspace{-3mm}
\label{table:cosal}
\end{table}

\begin{table*}[h]
\centering
\begin{tabular}{||c c c c c c c||} 
 \hline
 Model &  Precision@0.5 & Precision@0.6& Precision@0.7& Precision@0.8& Precision@0.9 &IoU   \\ [0.5ex] 
 \hline\hline
 SCRC bbox \cite{hu2016natural} & 9.73\% & 4.43\% & 1.51\% & 0.27\% & 0.03\% & 21.72\% \\
 GroundeR bbox \cite{donahue2015long} & 11.08\% &6.20\% &2.74\% &0.78\% &0.20\% &20.50\%
 \\
 Hu, etc.\cite{hu2016segmentation} & \textbf{34.02}\% & 26.71\% & \textbf{19.32}\% & 11.63\% & 3.92\% & 48.03\% \\  
 Our model & 32.53\% & \textbf{27.9}\% & 18.76\% & \textbf{12.37}\% & \textbf{4.37}\% & \textbf{50.09}\%   \\ 
 \hline
\end{tabular}
\caption{The results of previous models and our model on the ReferIt dataset.}
\vspace{-3mm}
\label{table:referit1}
\end{table*}

\vspace{-5mm}
\paragraph{Metric} For this task, we use \textit{average IoU over nine shapes and the background} as the evaluation metric. Specifically, for each region, we compute the intersection-over-union (IoU), which is the ratio of the total intersection area to the total union area of predicted colors and ground truth colors. We also compute the IoU for the background (white) of each image. The IoU for $10$ classes ($9$ shapes $+$ $1$ background) are computed over the entire test set and then averaged. 
\vspace{-5mm}
\paragraph{Model Implementation} A six-layer convolutional network is implemented as the image feature extractor. Each layer has a $3\times 3$ kernel with stride $1$ and output dimension $4,4,8,8,16,16$. ReLU is used for nonlinearity after each layer, and a max-pooling layer with a kernel of size 2 is inserted after every two layers. Each sentence in the textual description is encoded with bidirectional LSTMs that share parameters. Another LSTM with attention is put on top of the encoded sentences. The LSTMs have $16$ units. In the fusion network, the attention model has $16$ units, the GRU cells use $16$ units, and the termination gate uses a linear map on top of the hidden state of each GRU cell. Two convolutional layers of kernel size $1\times 1$ with the output dimension of $16,7$ are put on top of the fused features as a classifier. Then an upsampling layer is implemented on top of it, with a single-layer deconvolutional network of kernel size $16$, stride $8$ to upsample the classifier to the original resolution. The upsampling layer is initialized with bilinear transforms. The maximum of termination steps $T$ vary from $1$ to $4$. When $T=1$, the model is reduced to simply concatenating features extracted from the convolutional network with the last vector from LSTM. 
\vspace{-5mm}
\paragraph{Results} 
Results in Table \ref{table:cosal} show that the model with attention and $T=4$ achieves a better performance when there are more relational descriptions in the dataset. When there are more direct descriptions, the two models achieve similar performance. 
This demonstrates the framework's capability of interpreting multiple-sentence descriptions and associating them with their source image.

Figure \ref{fig:cosal2} illustrates how the model with $T=3$ interprets the nine sentences during each inference step. In each step, we take the sentence with the largest attention score as the one being attended to. Sentences in red are attended to in the first step. Those in yellow and green are attended to in the next two consecutive steps. We observe that the model tends to first extract information from direct descriptions, and then extract information from relational descriptions via reasoning. 



\subsection{Experiments on ReferIt} 
\paragraph{Dataset} The ReferIt dataset is composed of $19,894$ photographs of real world scenes, along with $130,525$ natural language descriptions on $96,654$ distinct objects in those photographs \cite{KazemzadehOrdonezMattenBergEMNLP14}. The dataset contains $238$ different object categories, including animals, people, buildings, objects and background elements (e.g., grass, sky). Both training and development datasets include $10,000$ images.
\vspace{-5mm}
\paragraph{Metric} Following \cite{hu2016segmentation}, we use two metrics for evaluation: 1) \textit{overall intersection-over-union (overall IoU)} of the predicted and ground truth of each region, averaged over the entire test set; 2) \textit{precision@threshold}, the percentage of test data whose (per image) IoU between prediction and ground truth is above the threshold. 
Thresholds are set to $0.5,0.6,0.7,0.8,0.9$. 
\vspace{-5mm}
\paragraph{Model Implementation}
A VGG-16 model \cite{simonyan2014very} is used as the image encoder for images of size $512\times 512$. Textual descriptions are encoded with an LSTM of $1,024$ units. In the fusion network, the attention model uses $512$ units and the GRU cells $1,024$ units, on top of which is a classifier and an upsampling layer similar to the implementation in Section 4.1.
The maximum number of inference steps is $3$. ReLU is used on top of each convolutional layer. $L2$-normalization is applied to the parameters of the network. 
\vspace{-5mm}
\paragraph{Results}

Table \ref{table:referit1} shows the experimental results of our model and the previous methods on the ReferIt dataset. We see that our framework yields a better IoU and precision than \cite{hu2016segmentation}. We attribute the superior performance to the unique attention mechanism used by our fusion network. It efficiently associates individual descriptive sentences with different regions of the source image. There is not much discrepancy between the two models with $T=1$ and $T=3$, probably due to the fact that most textual descriptions in this dataset are simple. 

\subsection{Experiments on Oxford-102 Flower Dataset}
\begin{figure*}[h!]
\centering
\includegraphics[width=1.3cm, height=1.3cm]{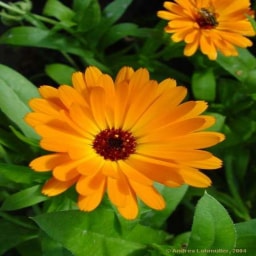} 
\includegraphics[width=1.3cm, height=1.3cm]{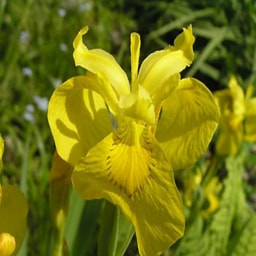} 
\includegraphics[width=1.3cm, height=1.3cm]{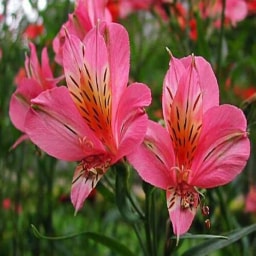} 
\includegraphics[width=1.3cm, height=1.3cm]{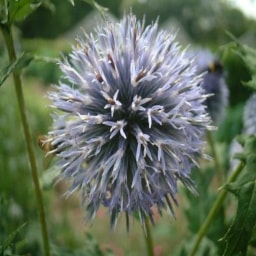} 
\includegraphics[width=1.3cm, height=1.3cm]{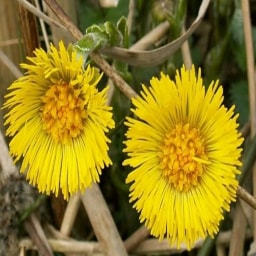} 
\includegraphics[width=1.3cm, height=1.3cm]{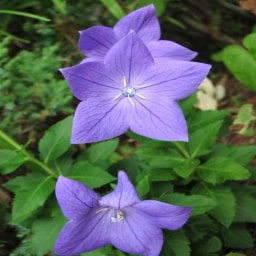} 
\includegraphics[width=1.3cm, height=1.3cm]{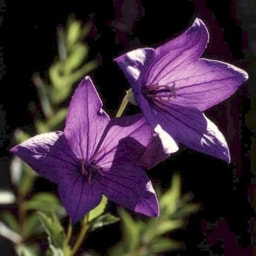} 
\includegraphics[width=1.3cm, height=1.3cm]{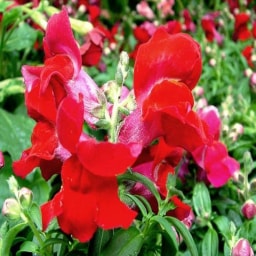} 
\includegraphics[width=1.3cm, height=1.3cm]{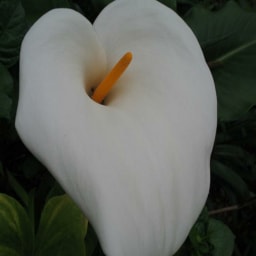} 
\includegraphics[width=1.3cm, height=1.3cm]{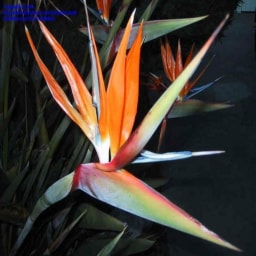} 
\includegraphics[width=1.3cm, height=1.3cm]{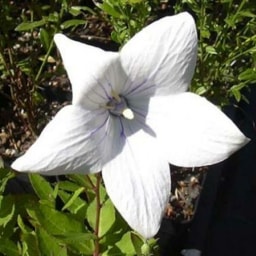}
\includegraphics[width=1.3cm, height=1.3cm]{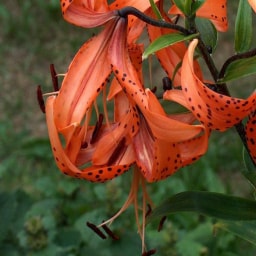} \\
\includegraphics[width=1.3cm, height=1.3cm]{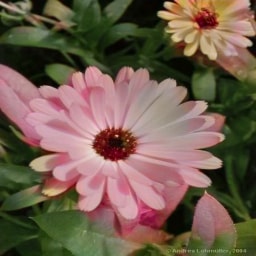} 
\includegraphics[width=1.3cm, height=1.3cm]{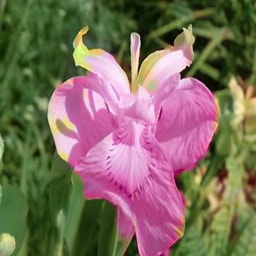} 
\includegraphics[width=1.3cm, height=1.3cm]{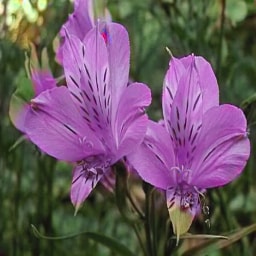} 
\includegraphics[width=1.3cm, height=1.3cm]{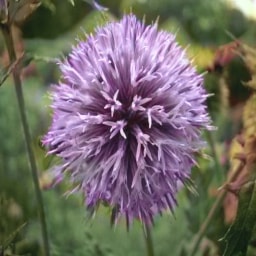} 
\includegraphics[width=1.3cm, height=1.3cm]{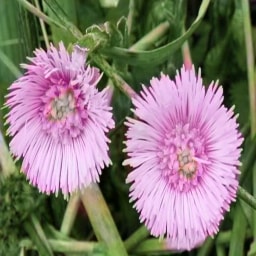} 
\includegraphics[width=1.3cm, height=1.3cm]{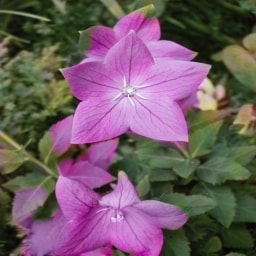} 
\includegraphics[width=1.3cm, height=1.3cm]{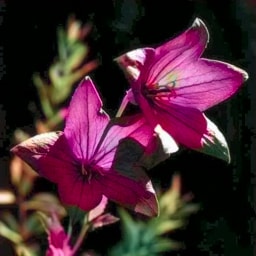} 
\includegraphics[width=1.3cm, height=1.3cm]{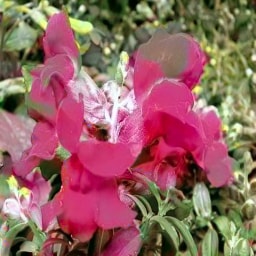} 
\includegraphics[width=1.3cm, height=1.3cm]{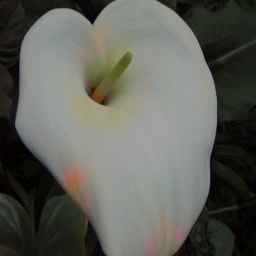} 
\includegraphics[width=1.3cm, height=1.3cm]{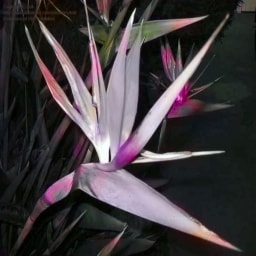} 
\includegraphics[width=1.3cm, height=1.3cm]{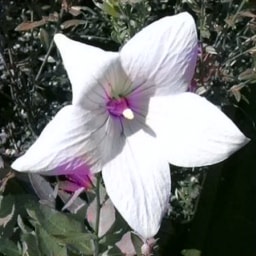}
\includegraphics[width=1.3cm, height=1.3cm]{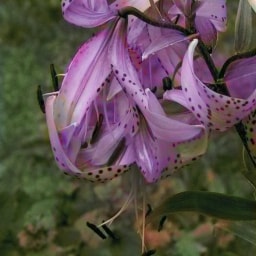}\\ 
\includegraphics[width=1.3cm, height=1.3cm]{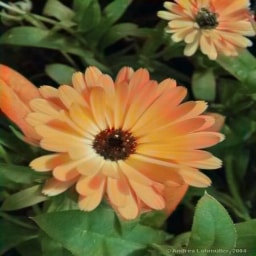} 
\includegraphics[width=1.3cm, height=1.3cm]{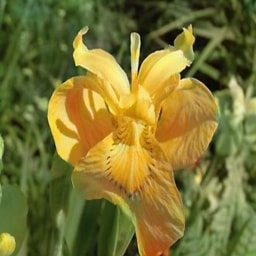}
\includegraphics[width=1.3cm, height=1.3cm]{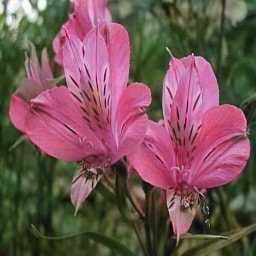} 
\includegraphics[width=1.3cm, height=1.3cm]{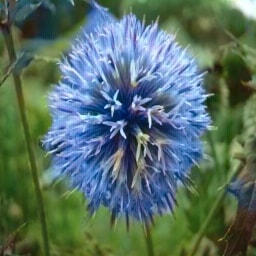}
\includegraphics[width=1.3cm, height=1.3cm]{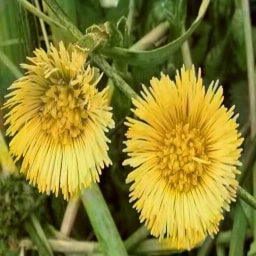} 
\includegraphics[width=1.3cm, height=1.3cm]{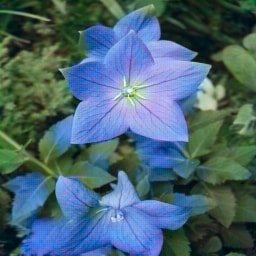}
\includegraphics[width=1.3cm, height=1.3cm]{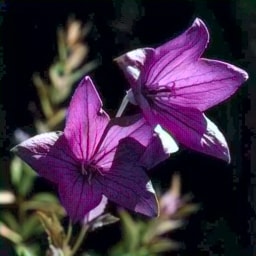} 
\includegraphics[width=1.3cm, height=1.3cm]{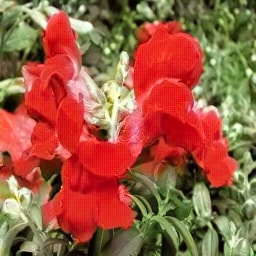}
\includegraphics[width=1.3cm, height=1.3cm]{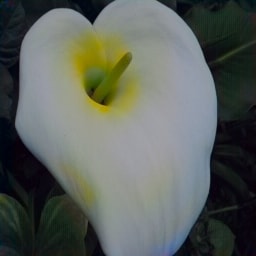} 
\includegraphics[width=1.3cm, height=1.3cm]{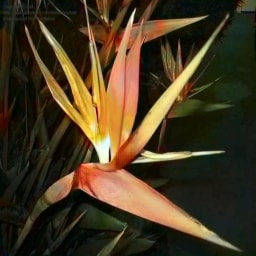}
\includegraphics[width=1.3cm, height=1.3cm]{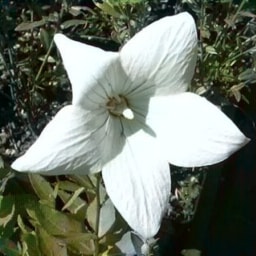} 
\includegraphics[width=1.3cm, height=1.3cm]{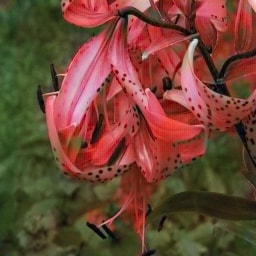}  
\caption{First row: original images. Second row: results from the image-to-image translation model in \cite{isola2016image}, without text input. Third row: results from our model, taking textual descriptions into account. The textual descriptions and more examples can be found in Appendix.}
\label{fig:flower1} 
\end{figure*}

\begin{figure*}[h!]
\centering
\includegraphics[width=1.3cm, height=1.3cm]{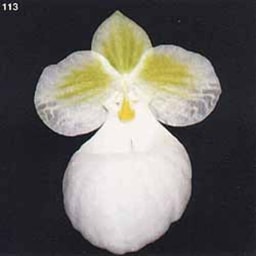}
\includegraphics[width=1.3cm, height=1.3cm]{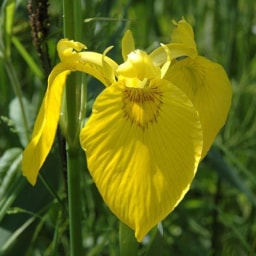} 
\includegraphics[width=1.3cm, height=1.3cm]{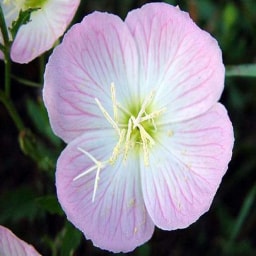} 
\includegraphics[width=1.3cm, height=1.3cm]{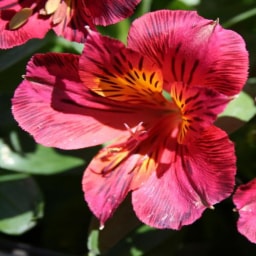} 
\includegraphics[width=1.3cm, height=1.3cm]{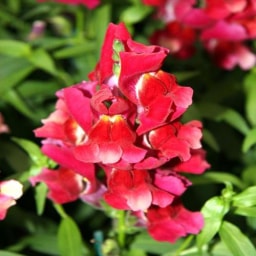} 
\includegraphics[width=1.3cm, height=1.3cm]{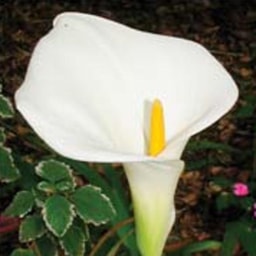} 
\includegraphics[width=1.3cm, height=1.3cm]{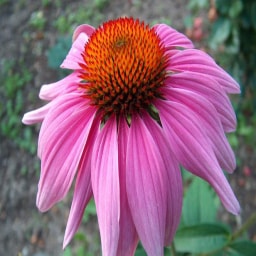} 
\includegraphics[width=1.3cm, height=1.3cm]{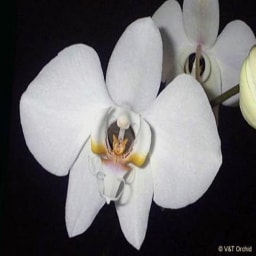} 
\includegraphics[width=1.3cm, height=1.3cm]{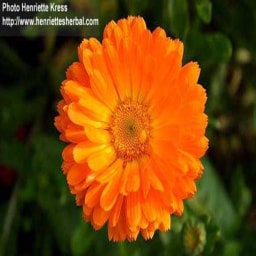} 
\includegraphics[width=1.3cm, height=1.3cm]{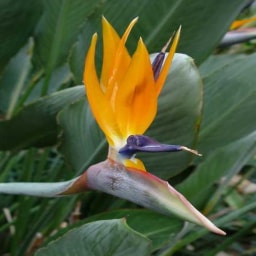} 
\includegraphics[width=1.3cm, height=1.3cm]{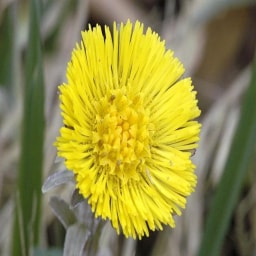} 
\includegraphics[width=1.3cm, height=1.3cm]{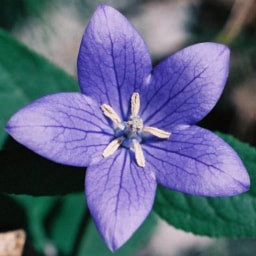} \\
\includegraphics[width=1.3cm, height=1.3cm]{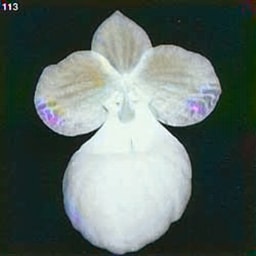}
\includegraphics[width=1.3cm, height=1.3cm]{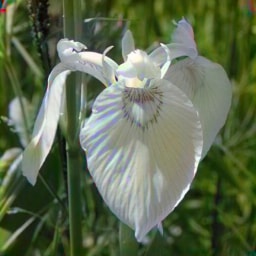}
\includegraphics[width=1.3cm, height=1.3cm]{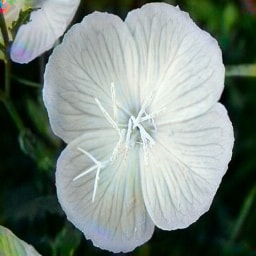}
\includegraphics[width=1.3cm, height=1.3cm]{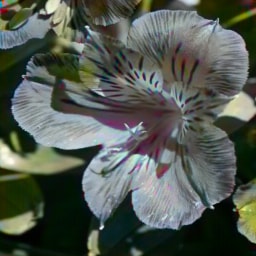}
\includegraphics[width=1.3cm, height=1.3cm]{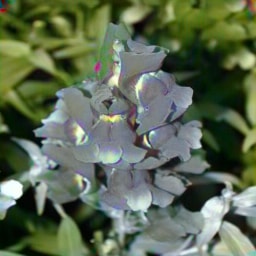}
\includegraphics[width=1.3cm, height=1.3cm]{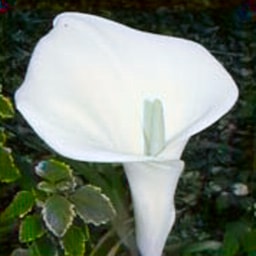}
\includegraphics[width=1.3cm, height=1.3cm]{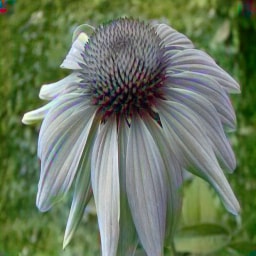}
\includegraphics[width=1.3cm, height=1.3cm]{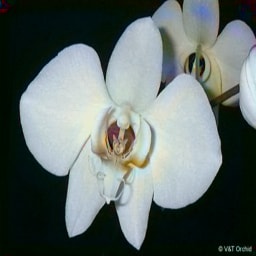}
\includegraphics[width=1.3cm, height=1.3cm]{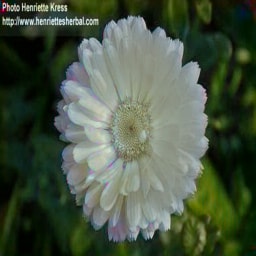}
\includegraphics[width=1.3cm, height=1.3cm]{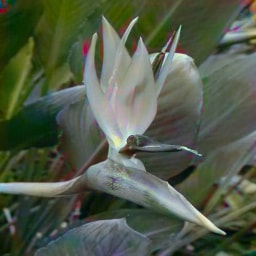}
\includegraphics[width=1.3cm, height=1.3cm]{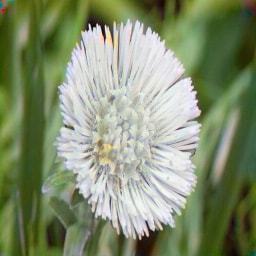}
\includegraphics[width=1.3cm, height=1.3cm]{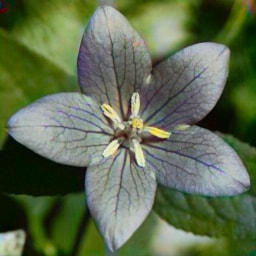} \\
\includegraphics[width=1.3cm, height=1.3cm]{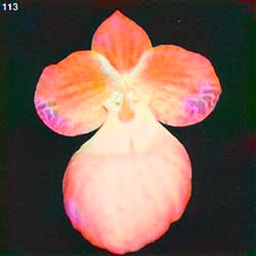}
\includegraphics[width=1.3cm, height=1.3cm]{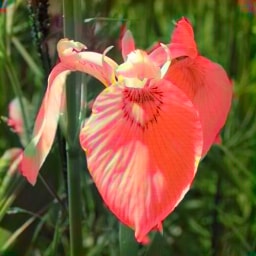}
\includegraphics[width=1.3cm, height=1.3cm]{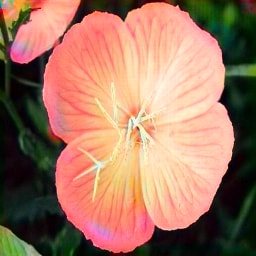}
\includegraphics[width=1.3cm, height=1.3cm]{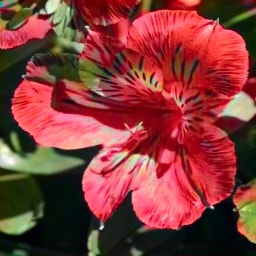}
\includegraphics[width=1.3cm, height=1.3cm]{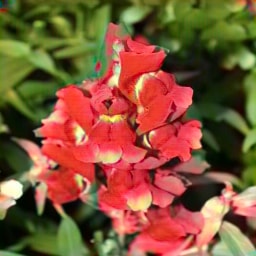}
\includegraphics[width=1.3cm, height=1.3cm]{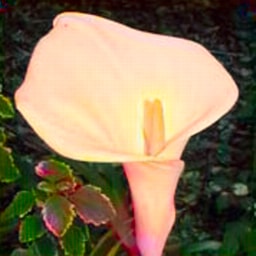}
\includegraphics[width=1.3cm, height=1.3cm]{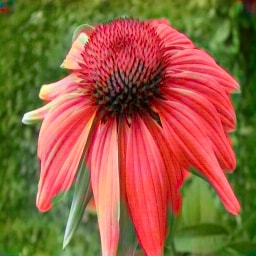}
\includegraphics[width=1.3cm, height=1.3cm]{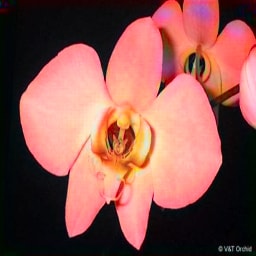}
\includegraphics[width=1.3cm, height=1.3cm]{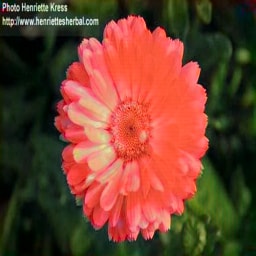}
\includegraphics[width=1.3cm, height=1.3cm]{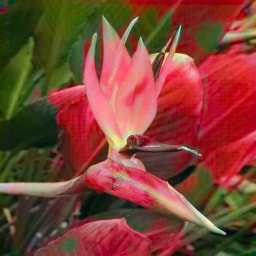}
\includegraphics[width=1.3cm, height=1.3cm]{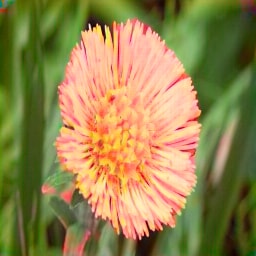}
\includegraphics[width=1.3cm, height=1.3cm]{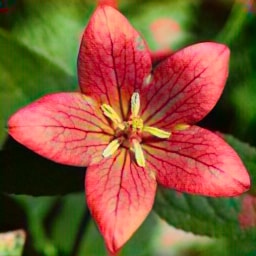} \\
\includegraphics[width=1.3cm, height=1.3cm]{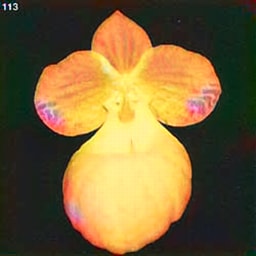}
\includegraphics[width=1.3cm, height=1.3cm]{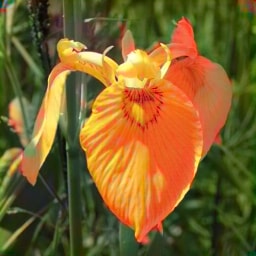}
\includegraphics[width=1.3cm, height=1.3cm]{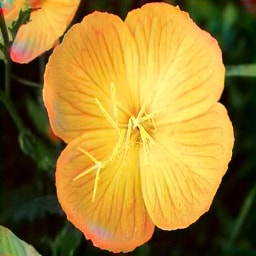}
\includegraphics[width=1.3cm, height=1.3cm]{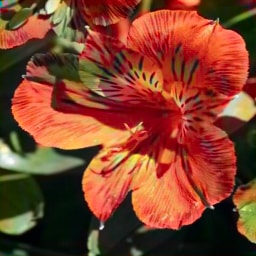}
\includegraphics[width=1.3cm, height=1.3cm]{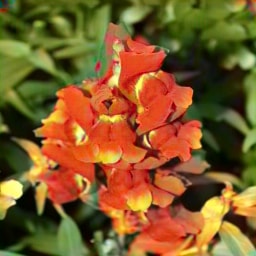}
\includegraphics[width=1.3cm, height=1.3cm]{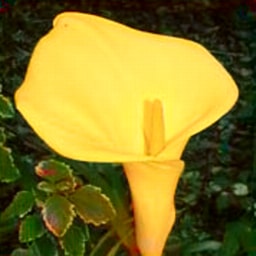}
\includegraphics[width=1.3cm, height=1.3cm]{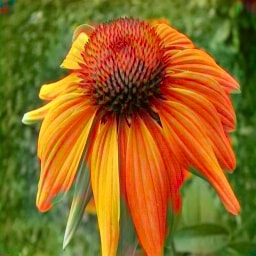}
\includegraphics[width=1.3cm, height=1.3cm]{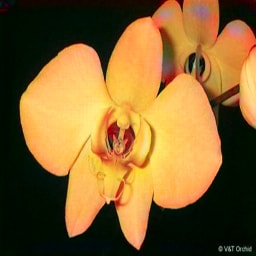}
\includegraphics[width=1.3cm, height=1.3cm]{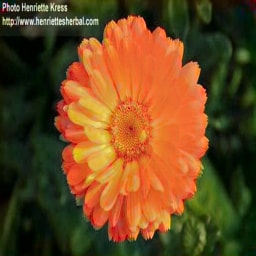}
\includegraphics[width=1.3cm, height=1.3cm]{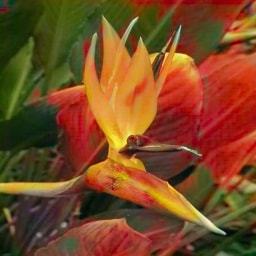}
\includegraphics[width=1.3cm, height=1.3cm]{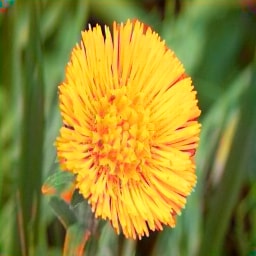}
\includegraphics[width=1.3cm, height=1.3cm]{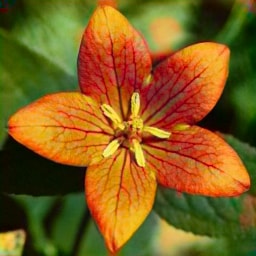} \\
\includegraphics[width=1.3cm, height=1.3cm]{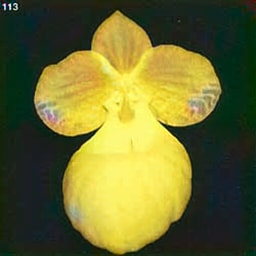}
\includegraphics[width=1.3cm, height=1.3cm]{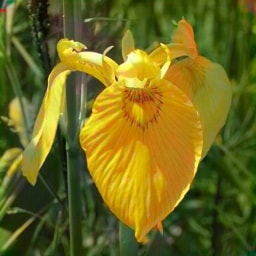}
\includegraphics[width=1.3cm, height=1.3cm]{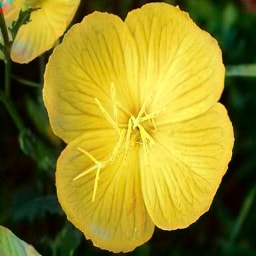}
\includegraphics[width=1.3cm, height=1.3cm]{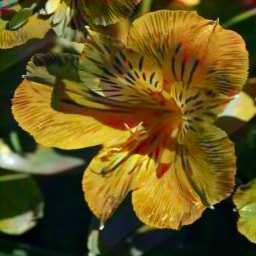}
\includegraphics[width=1.3cm, height=1.3cm]{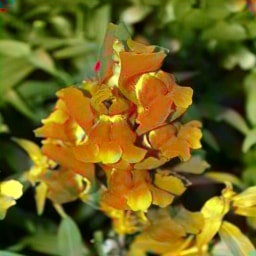}
\includegraphics[width=1.3cm, height=1.3cm]{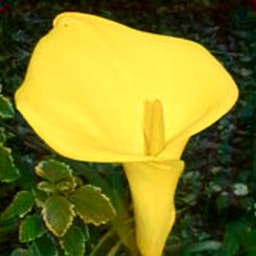}
\includegraphics[width=1.3cm, height=1.3cm]{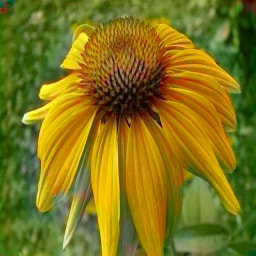}
\includegraphics[width=1.3cm, height=1.3cm]{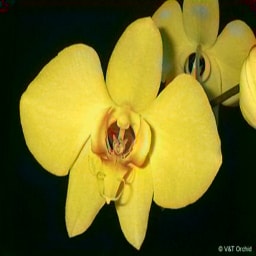}
\includegraphics[width=1.3cm, height=1.3cm]{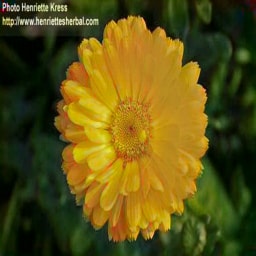}
\includegraphics[width=1.3cm, height=1.3cm]{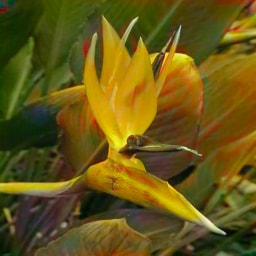}
\includegraphics[width=1.3cm, height=1.3cm]{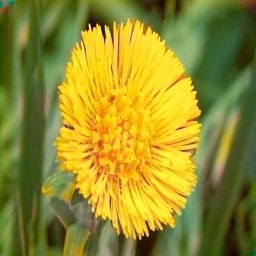}
\includegraphics[width=1.3cm, height=1.3cm]{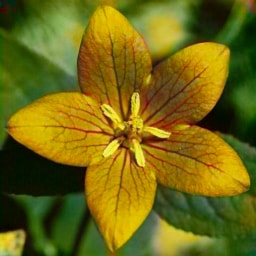} \\ 
\includegraphics[width=1.3cm, height=1.3cm]{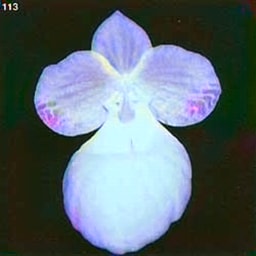}
\includegraphics[width=1.3cm, height=1.3cm]{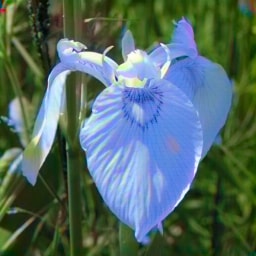}
\includegraphics[width=1.3cm, height=1.3cm]{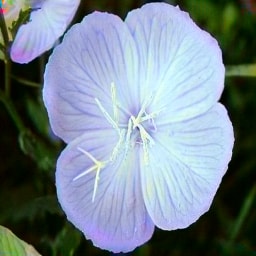}
\includegraphics[width=1.3cm, height=1.3cm]{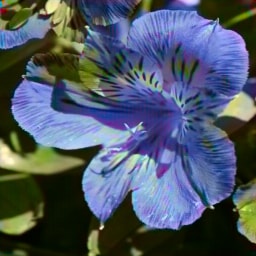}
\includegraphics[width=1.3cm, height=1.3cm]{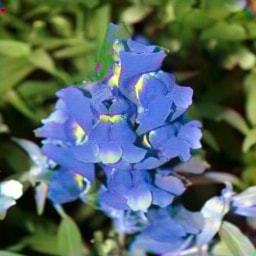}
\includegraphics[width=1.3cm, height=1.3cm]{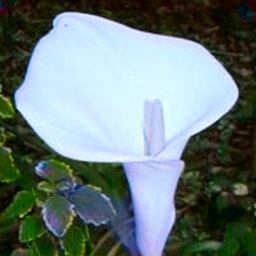}
\includegraphics[width=1.3cm, height=1.3cm]{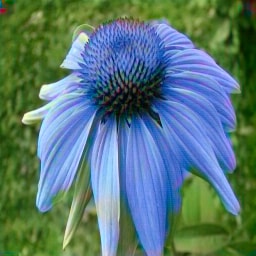}
\includegraphics[width=1.3cm, height=1.3cm]{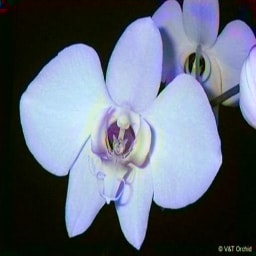}
\includegraphics[width=1.3cm, height=1.3cm]{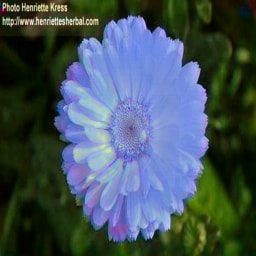}
\includegraphics[width=1.3cm, height=1.3cm]{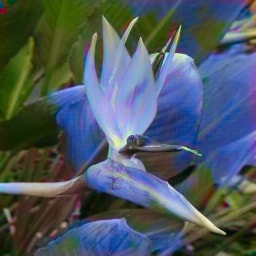}
\includegraphics[width=1.3cm, height=1.3cm]{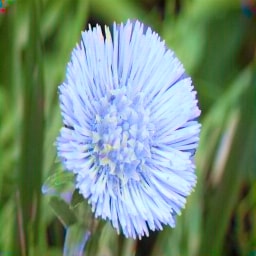}
\includegraphics[width=1.3cm, height=1.3cm]{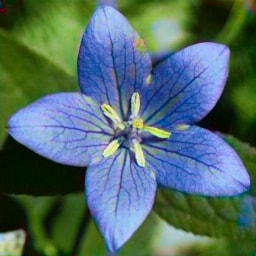} \\
\includegraphics[width=1.3cm, height=1.3cm]{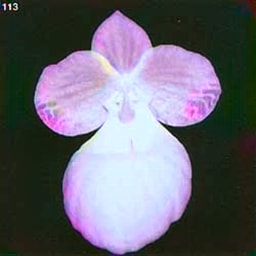}
\includegraphics[width=1.3cm, height=1.3cm]{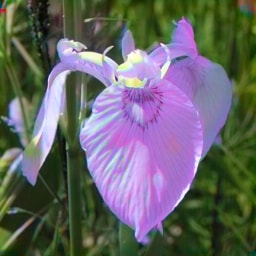}
\includegraphics[width=1.3cm, height=1.3cm]{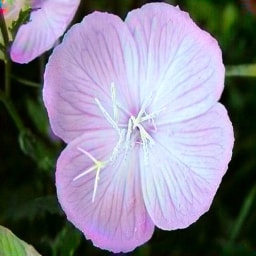}
\includegraphics[width=1.3cm, height=1.3cm]{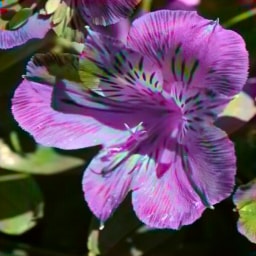}
\includegraphics[width=1.3cm, height=1.3cm]{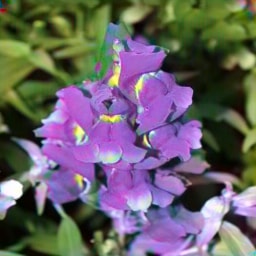}
\includegraphics[width=1.3cm, height=1.3cm]{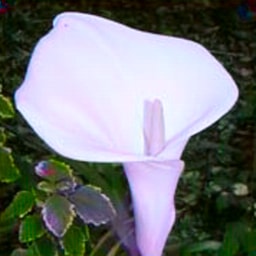}
\includegraphics[width=1.3cm, height=1.3cm]{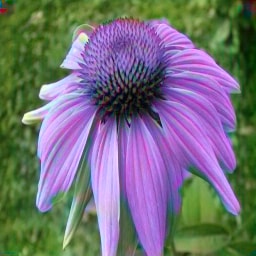}
\includegraphics[width=1.3cm, height=1.3cm]{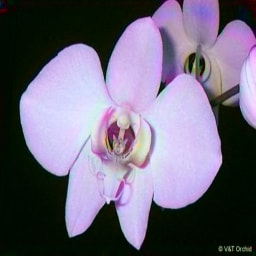}
\includegraphics[width=1.3cm, height=1.3cm]{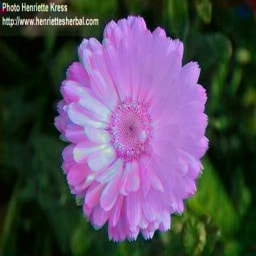}
\includegraphics[width=1.3cm, height=1.3cm]{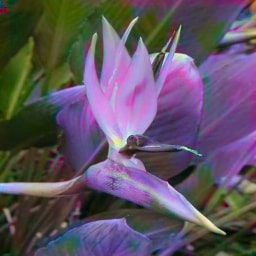}
\includegraphics[width=1.3cm, height=1.3cm]{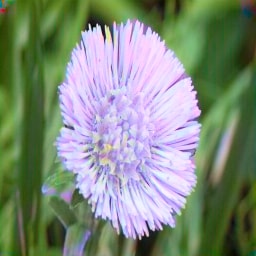}
\includegraphics[width=1.3cm, height=1.3cm]{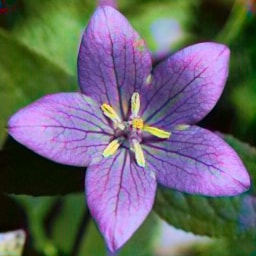} 
\caption{First row: original images. Remaining rows: results generated from our framework with arbitrary text input: \textit{``The flower is white/red/orange/yellow/blue/purple in color''}. }
\label{fig:flower2} 
\end{figure*}
\paragraph{Dataset} 
The Oxford-102 Flowers dataset \cite{Nilsback08} contains $8,189$ images from $102$ flower categories. Each image has five textual descriptions \cite{reed2016generative}. Following \cite{reed2016generative}, \cite{reed2016learning} and \cite{akata2015evaluation}, we split the dataset into 82 classes for training and 20 classes for testing. The task is defined as follows: Given a grayscale image of a flower and a description of the shapes and colors of the flower, colorize the image according to the description.

\vspace{-2mm}
\paragraph{Model Implementation}
A 15-layer convolutional network similar to \cite{zhang2016colorful} is used for encoding $256\times 256$ images. Textual descriptions are encoded with an bidirectional LSTM of $512$ units. In the fusion network, the attention model uses $128$ units and the GRU cells $128$ units. The image encoder 
is composed of $2$ deconvolutional layers, each followed by $2$ convolutional layers, to upsample the fusion feature map to the target image space of $256\times 256\times 2$. The maximum length of the spatial RNN is $1$. The discriminator is composed of 5 layers of convolutional networks of stride $2$, with the output dimension $256,128,64,32,31$. The discriminator score is the average of the final output. ReLU is used for nonlinearity following each convolutional layer, except for the last one which uses the sigmoid function.  
\vspace{-2mm}
\paragraph{Setup} Due to the lack of available models for the task, we compare our framework with a previous model developed for image-to-image translation as baseline, which colorizes images without text descriptions. We carried out two human evaluations using Mechanical Turk to compare the performance of our model and the baseline. For each experiment, we randomly sampled 1,000 images from the test set and then turned these images into black and white. For each image, we generated a pair of two images using our model and the baseline, respectively. Our model took into account the caption in generation while the baseline did not. Then we randomly permuted the 2,000 generated images. In the first experiment, we presented to human annotators the 2,000 images, together with their original captions, and asked humans to rate the consistency between the generated images and the captions in a scale of $0$ and $1$, with $0$ indicating no consistency and $1$ indicating consistency. In the second experiment, we presented to human annotators the same 2,000 images without captions, but asked human annotators to rate the quality of each image without providing its original caption. The quality was rated in a scale of $0$ and $1$, with $0$ indicating low quality and $1$ indicating high quality. 


\vspace{-2mm}
\paragraph{Results} The results of comparison are shown in Table~\ref{table:flower}. Our model achieves better consistency with captions and also better image quality by making use of information in captions. The colorization results on 10 randomly-sampled images from the test set are shown in Figure \ref{fig:flower1}. As we can see, without text input, the baseline approach often colorizes images with the same color (in this dataset, most images are painted with purple, red or white), while our framework can generate flowers similar to their original colors which are specified in texts. Figure \ref{fig:flower2} provides some example images generated with arbitrary text description using our model.

\vspace{-1mm}
\begin{table}[H]
\centering
\begin{tabular}{||c c c c||} 
 \hline
  &  Our Model & BaseLine & Truth\\ [0.5ex] 
 \hline\hline
Consistency & \textbf{0.849} & 0.27 & N/A \\
\hline
Quality & 0.598 & 0.404 & \textbf{0.856}\\ 
 \hline
\end{tabular}
\caption{The average rate of consistency with captions and image quality for our model and the baseline model respectively, averaged over $1,000$ images. The average quality of 1,000 truth images from the data set is also provided for comparison.}
\label{table:flower}
\vspace{-2mm}
\end{table}

\vspace{-2mm}
\section{Conclusion and Future Work}
In this paper we introduce the problem of Language-Based Image Editing (LBIE), and propose a generic modeling framework for two sub-tasks of LBIE: language-based image segmentation and colorization. At the heart of the proposed framework is a fusion module that uses recurrent attentive models to dynamically decide, for each region of an image, whether to continue the text-to-image fusion process. Our models have demonstrated superior empirical results on three datasets: the ReferIt dataset for image segmentation, the Oxford-102 Flower dataset for colorization, and the synthetic CoSaL dataset for evaluating the end-to-end performance of the LBIE system. In future, we will extend the framework to other image editing subtasks and build a dialogue-based image editing system that allows users to edit images interactively.



{\small
\bibliographystyle{ieee}
\bibliography{paper}
}
\clearpage
\onecolumn
\newpage 
\appendix
\section{Captions for Figure 6}   
\begin{table*}[h!]
\centering
\begin{tabular}{||c| c | c| c| p{8cm}||} 
\hline 
 \hline
 B\&W & Original &I2I \cite{isola2016image} & LBIE & Caption  \\ [0.5ex] 
 \hline   
\figsixtable{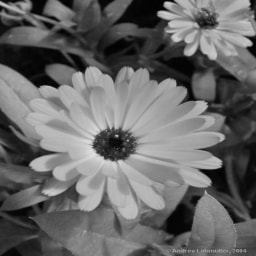}{this flower has petals that are orange with red stamen}
 \figsixtable{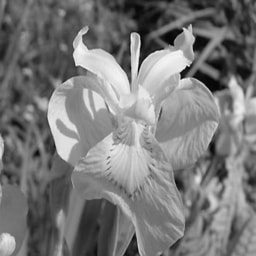}{the petals on this flower are yellow with a few brown dots}
 \figsixtable{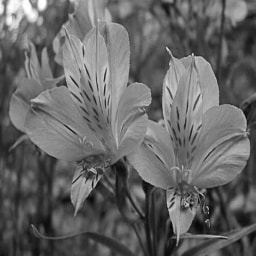}{this flower has pink petals containing brown  yellow spots}
 \figsixtable{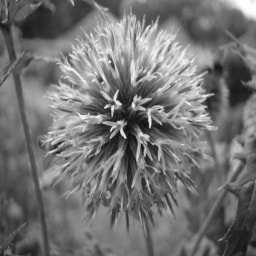}{this flower has large number light blue tubular petals arranged in globe configuration}
 \figsixtable{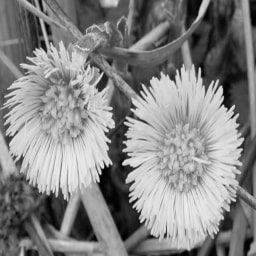}{the petals of this flower are yellow and the background is green}
 \figsixtable{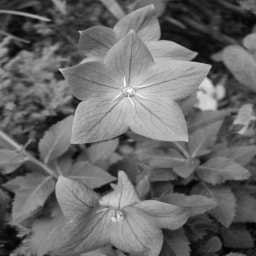}{the blue flower is star shaped and it has petal that is soft. it has stamens in the middle.}
 \figsixtable{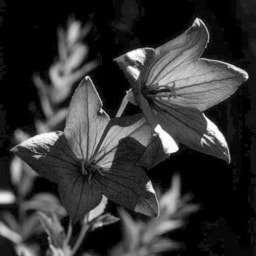}{this flower has dark purple petals with light purple anther filaments stamen.}
 \figsixtable{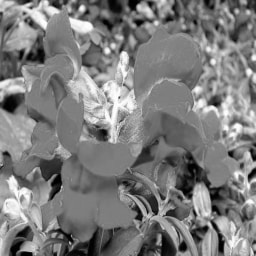}{this flower has large wide petals in brilliant red}
 \figsixtable{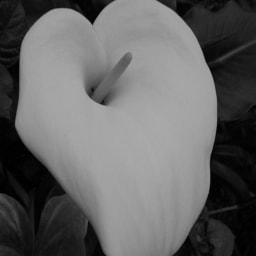}{this flower has petals that are white with yellow style}
 \figsixtable{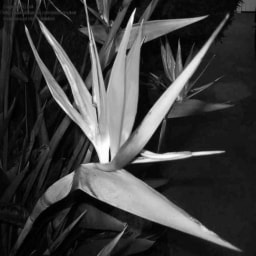}{this flower is made up of long pointy orange petals that surround the dark purple pistil}
 \figsixtable{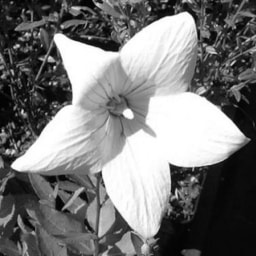}{the petals of this flower are white with a long stigma}
 \figsixtable{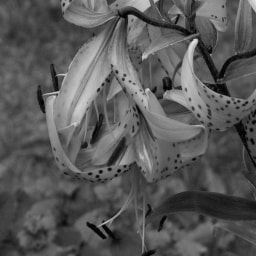}{this red flower is spotted and has very large filament}
\end{tabular}
\caption{The five columns contain input black and white images, original color images, images generated without captions, images generated with captions, and the captions respectively.}
\label{table:supp1}
\end{table*}

\newpage
\section{More examples on the Oxford-102 Flowers dataset}
\begin{table*}[h!]
\centering
\begin{tabular}{||c| c | c| c| p{8cm}||} 
\hline 
 \hline
 B\&W&Original &I2I \cite{isola2016image} & LBIE & Caption  \\ [0.5ex] 
 \hline 
\figsupptable{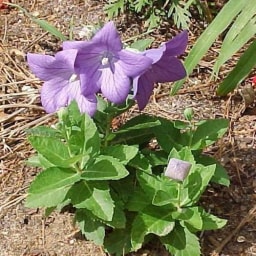}{this particular flower has petals that are green  purple} 
\figsupptable{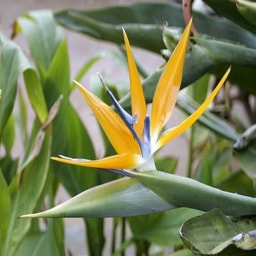}{this flower has yellow petals as well as  green  } 
\figsupptable{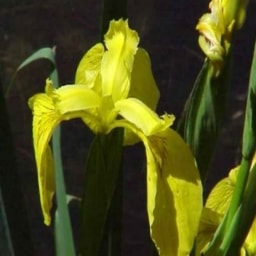}{this flower is yellow in color  with petals that are droopy} 
\figsupptable{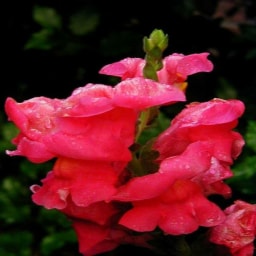}{this flower has bright pink petals in rows extending down  dark green pistil } 
\figsupptable{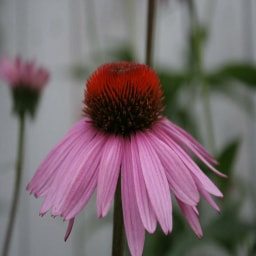}{this flower has long slender petals that are pink that drop down surrounding  large red pompom center} 
\figsupptable{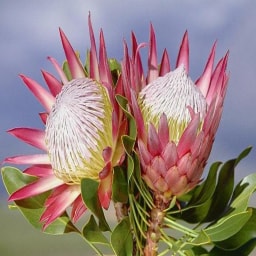}{this cone shaped flower has pink petals  many yellow  white stamen that form  large protruding cylinder in the center} 
\figsupptable{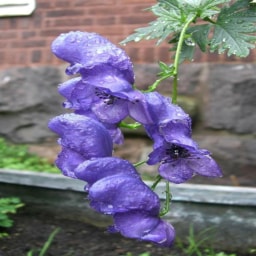}{this flower consist   bunch  purple pedals climbing down the stock with black stamen} 
\figsupptable{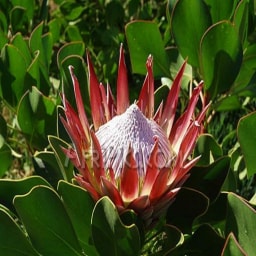}{this flower has petals that are purple  has red points} 
\figsupptable{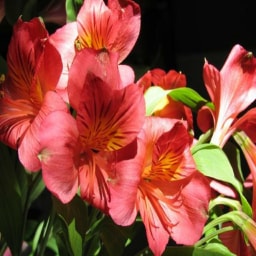}{this flower has  red petal   red stamen in the center} 
\figsupptable{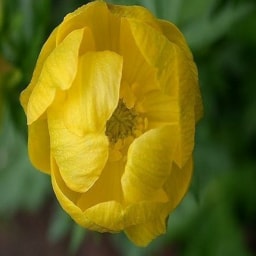}{this flower is yellow in color  with petals that are curled inward} 
\figsupptable{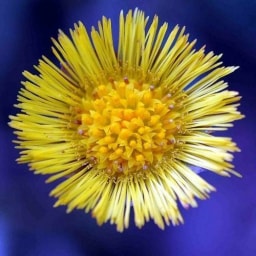}{the flower has lots  tiny yellow petals surround yellow stamen in the middle} 
\figsupptable{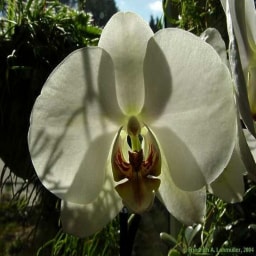}{this odd flower has five large petals with  ovule   receptacle in the center} 
\figsupptable{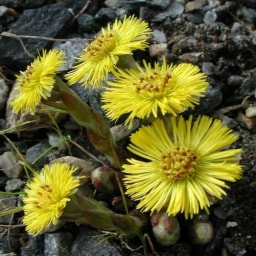}{the flower is yellow with petals that are needle like  soft  smooth  separately arranged in  disc like manner around stamens} 
\figsupptable{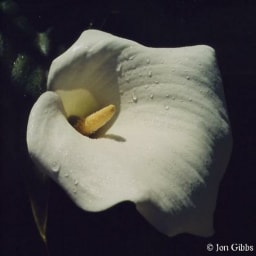}{this flower is white in color  with only one large petal} 
\figsupptable{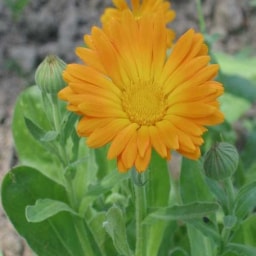}{this flower has petals that are orange  has yellow stamen} 
\figsupptable{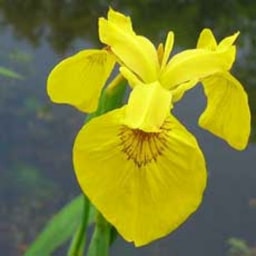}{this flower is yellow in color   has petals that are oval shaped  drooping} 
\end{tabular}
\caption{The five columns contain input black and white images, original color images, images generated without captions, images generated with captions, and the captions respectively.}
\label{table:supp2}
\end{table*}

\begin{table*}[h!]
\centering
\begin{tabular}{||c| c | c| c| p{8cm}||} 
\hline 
 \hline
\figsupptable{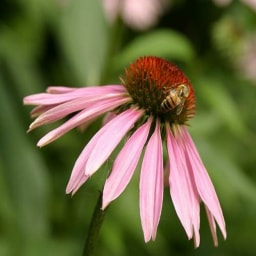}{spiky reddish anthers atop greenish stamen  long  thin pinkish purple petals} 
\figsupptable{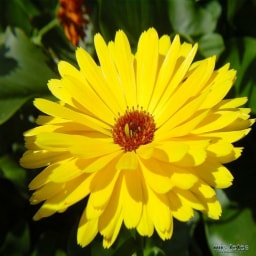}{the flower has many yellow petals surround the red stamen} 
\figsupptable{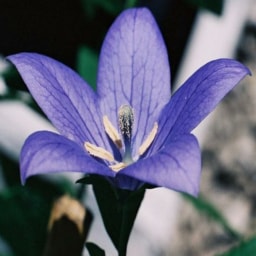}{flower with violet petals that have long  large stamen at it  center} 
\figsupptable{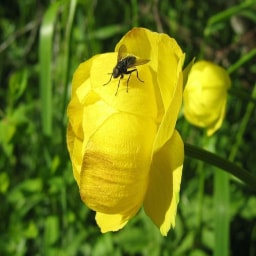}{this flower has petals that are yellow  folded together}  
\figsupptable{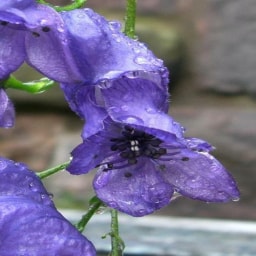}{these flowers have purple  lavender petals with black stamen  green}

\figsupptable{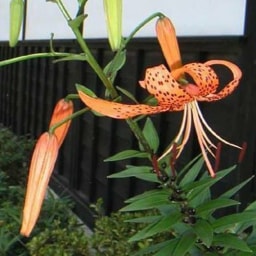}{this flower has large orange petals with long white stamens }  
\figsupptable{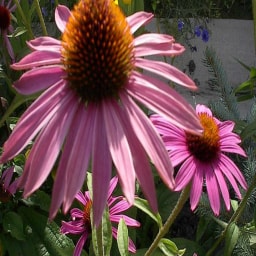}{what i like about this flower is its acorn looking stigma}  
\figsupptable{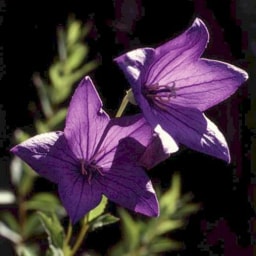}{this flower has dark purple petals with light purple anther filaments  stamen }  
\figsupptable{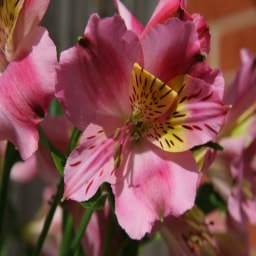}{this flower has multicolored petals that are pink with darker lines  yellow sections}  
\figsupptable{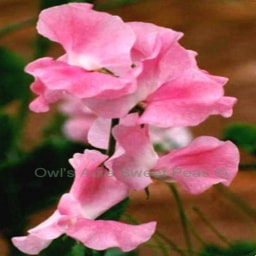}{this flower is pink in color   has petals that are wavy  thin }  
\figsupptable{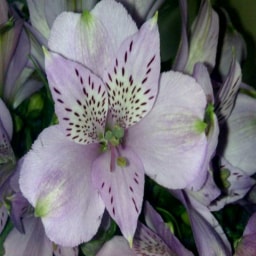}{this flower is purple  white in color  with petals that are spotted }  
\figsupptable{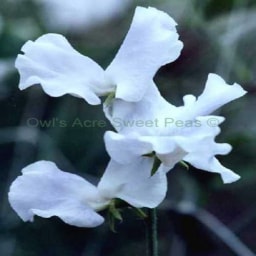}{this flower only has  few white petals but does have  visible green sepal }  
\figsupptable{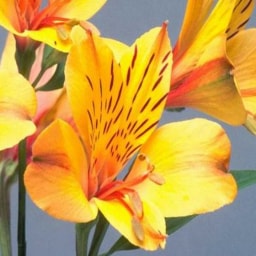}{the petals  the flower are yellow in color with orange accents  brown specks }  
\figsupptable{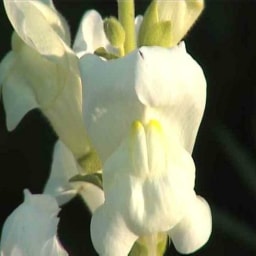}{this small puffy flower is white with  yellow tint at the tip}  
\figsupptable{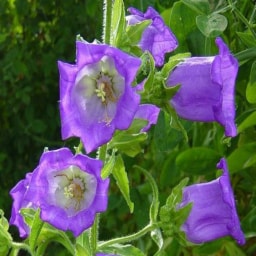}{this flower has large green sepals holding its purple bell shaped petals}   
\end{tabular}
\caption{(Cont.) The five columns contain input black and white images, original color images, images generated without captions, images generated with captions, and the captions respectively.}
\label{table:supp3}
\end{table*}

\end{document}